\documentclass[letterpaper]{article} 
\usepackage{aaai2026} 

\usepackage{times}  
\usepackage{helvet}  
\usepackage{courier}  
\usepackage[hyphens]{url}  
\usepackage{graphicx} 

\usepackage{multirow} 
\usepackage{array} 
\usepackage{natbib}  
\usepackage{caption} 
\usepackage{algorithm}
\usepackage{amsmath, amssymb}
\usepackage{tcolorbox}
\renewcommand{\arraystretch}{1.2} 
\usepackage{booktabs}

\usepackage{graphicx}
\usepackage{subcaption}

\usepackage[noend]{algpseudocode}
\usepackage{graphicx} 
\usepackage{amsmath,amssymb}     
\usepackage{amsthm}              
\usepackage{algpseudocode}       
\usepackage{microtype}           

\usepackage{caption}
\usepackage{booktabs}
\usepackage{siunitx}
\usepackage{newfloat}
\usepackage{listings}
\DeclareCaptionStyle{ruled}{labelfont=normalfont,labelsep=colon,strut=off} 
\floatstyle{ruled}
\newfloat{listing}{tb}{lst}{}
\floatname{listing}{Listing}
\title{DRIVE: Dynamic Rule Inference and Verified Evaluation \\ for Constraint-Aware Autonomous Driving} 

\author {
    Longling Geng\textsuperscript{\rm 1},
    Huangxing Li\textsuperscript{\rm 2},
    Viktor Lado Naess\textsuperscript{\rm 1},
    Mert Pilanci\textsuperscript{\rm 1},
}
\affiliations {
    \textsuperscript{\rm 1}Stanford University\\
    \textsuperscript{\rm 2}Microsoft\\
    gll2027@stanford.edu, huangxingli@microsoft.com, viktorln@stanford.edu, pilanci@stanford.edu
}

\begin{document}

\maketitle
\begin{abstract}
Understanding and adhering to soft constraints is essential for safe and socially compliant autonomous driving. However, such constraints are often implicit, context-dependent, and difficult to specify explicitly. In this work, we present DRIVE, a novel framework for Dynamic Rule Inference and Verified Evaluation that models and evaluates human-like driving constraints from expert demonstrations. DRIVE leverages exponential-family likelihood modeling to estimate the feasibility of state transitions, constructing a probabilistic representation of soft behavioral rules that vary across driving contexts. These learned rule distributions are then embedded into a convex optimization-based planning module, enabling the generation of trajectories that are not only dynamically feasible but also compliant with inferred human preferences. Unlike prior approaches that rely on fixed constraint forms or purely reward-based modeling, DRIVE offers a unified framework that tightly couples rule inference with trajectory-level decision-making. It supports both data-driven constraint generalization and principled feasibility verification. We validate DRIVE on large-scale naturalistic driving datasets, including inD, highD, and RoundD, and benchmark it against representative inverse constraint learning and planning baselines. Experimental results show that DRIVE achieves 0.0\% soft constraint violation rates, smoother trajectories, and stronger generalization across diverse driving scenarios. Verified evaluations further demonstrate the efficiency, explanability, and robustness of the framework for real-world deployment.
\end{abstract}

{\small
\begin{links}
    \link{Code}{https://github.com/genglongling/DRIVE} \\
    \link{Datasets}{https://github.com/genglongling/DRIVE}
\end{links}
}

\section{Introduction}

Imitating human driving behavior is a promising approach to enhance the safety and adaptability of autonomous vehicle (AV) decision-making~\cite{wang2021towards}. A key aspect of such behavior involves soft constraints, including comfort preferences, cautious reactions to uncertainty, and social norms. These constraints are often implicit, context-dependent, and difficult to encode through rule-based systems~\cite{huang2024general}.

While imitation learning and inverse reinforcement learning (IRL) aim to recover reward functions that explain expert behavior~\cite{abbeel2004apprenticeship}, they typically assume that all preferences can be captured through scalar reward optimization. This overlooks many feasibility-related constraints that are structural or latent in nature. Inverse constrained reinforcement learning (ICRL) addresses this issue by learning constraint functions alongside rewards~\cite{liu2024comprehensive, deshpande2025advances}, yet existing methods often adopt deterministic or handcrafted constraint forms, limiting their ability to represent the probabilistic and context-sensitive characteristics of human driving~\cite{gaurav2022learning}.

To address these limitations, we propose DRIVE, a unified framework for dynamic rule inference and verified evaluation. DRIVE uses exponential-family likelihood modeling~\cite{salhotra2022learning} to estimate the feasibility of state transitions from expert demonstrations, constructing probabilistic soft constraints instead of relying on binary specifications. These constraints are then incorporated into a convex trajectory optimization procedure, enabling the generation of plans that satisfy safety requirements while aligning with human preferences. By integrating learning, planning, and evaluation, DRIVE supports scalable and interpretable decision-making across diverse driving scenarios.

\section{Related Work}

\textbf{Imitation learning (IL) and inverse reinforcement learning (IRL).} IL and IRL aim to reproduce expert behavior by learning policies or reward functions from demonstrations~\cite{abbeel2004apprenticeship, zare2024survey}. In autonomous driving, these approaches have been used to model preferences related to safety, efficiency, and social compliance~\cite{li2025tw}. Maximum Entropy IRL~\cite{snoswell2020revisiting} introduces a stochastic policy formulation to encourage diverse behavior generation, while deep IRL methods enhance scalability and expressiveness in high-dimensional environments. However, most approaches reduce learning to reward maximization, assuming all constraints can be embedded in a scalar objective. This limits their ability to represent non-reward-driven rules, such as comfort margins or context-sensitive hesitation.

\textbf{Constrained reinforcement learning (CRL) and inverse constrained learning(ICL).} CRL incorporates safety or feasibility constraints into policy optimization, either by restricting action spaces or penalizing unsafe behaviors~\cite{yao2023constraint}. CMDP-based optimization~\cite{so2024solving} is commonly used for enforcing critical constraints in simulation and real-world driving. Inverse constrained reinforcement learning (ICRL) instead learns constraints from expert demonstrations~\cite{qiao2023multi, liu2022benchmarking}. Existing methods include learning polyhedral or linear forms~\cite{berkenkamp2019safe}, estimating support sets~\cite{brauckmann2021reinforcement}, and inferring rule masks or safe sets from data~\cite{saraswat2024revolutionizing}. However, most rely on binary or deterministic forms, which do not capture the uncertainty and smoothness of soft human preferences.

\textbf{Probabilistic constraint modeling and likelihood-based inference.} To overcome the limitations of deterministic approaches, studies have explored probabilistic formulations for constraint inference. Bayesian IRL~\cite{papadimitriou2024bayesian} models uncertainty in reward functions and reflects soft behavioral tendencies. Probabilistic ICRL methods, such as maximum likelihood constraint inference~\cite{subramanian2024confidence} and exponential-family soft constraint learning~\cite{shah2021computationally, gunasekar2014exponential}, express transition feasibility through likelihoods or posterior densities. Despite their flexibility, these often decouple constraint inference from planning or assume simplified environments.

\textbf{Trajectory optimization and learning-based planning.}  
Trajectory planning for AVs typically handles constraints via cost functions, feasibility filters, or model-based control. Model predictive control (MPC) applies physics-based constraints for short-horizon planning~\cite{di2018real}, while beam search and sampling-based methods prune infeasible options during search~\cite{schmid2020efficient}. Learning-augmented planners use neural cost models or policies to guide generation~\cite{khanal2025learning}, but often rely on manually defined rules and lack soft constraint learning. Generative models and multimodal IRL~\cite{xu2024integration} improve diversity, yet depend on fixed latent structures that are hard to interpret or misaligned with human reasoning.

While progress has been made in modeling expert behavior via rewards, constraints, and probabilistic inference, key challenges remain. Many IRL and ICRL methods view rewards and constraints as separate explanations, limiting expressiveness. Deterministic constraints often fail to capture the stochastic and context-aware nature of real driving. Probabilistic models, though more flexible, are seldom integrated with scalable planning frameworks. These gaps highlight the need for unified approaches that learn probabilistic soft rules and enable interpretable, scalable planning.

\section{Framework}

\begin{figure*}[htbp]
\centering
\includegraphics[width=0.89\textwidth]{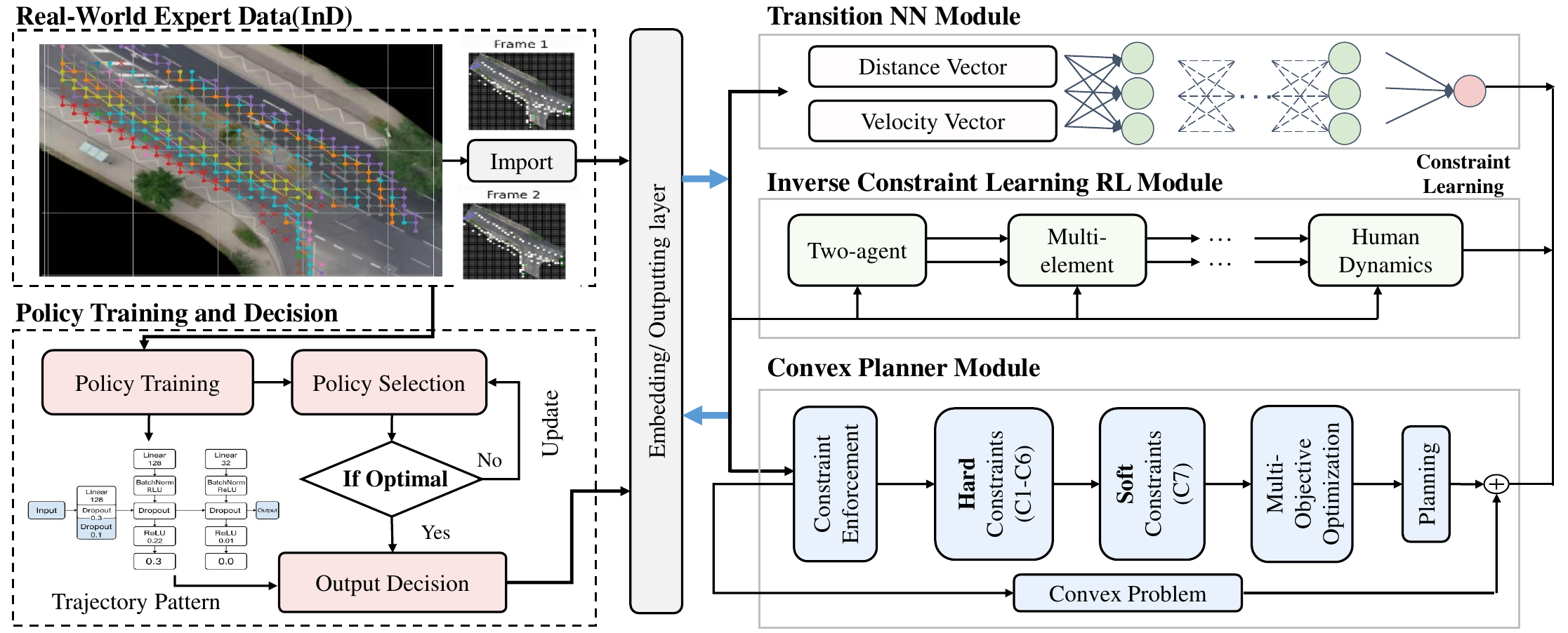}
\caption{The proposed DRIVE framework.}
\label{fig:drive_framework}
\end{figure*}

We propose DRIVE (Dynamic Rule Inference and Verified Evaluation), a unified framework that learns probabilistic soft constraints from expert driving demonstrations and integrates them into trajectory-level planning for safe and interpretable autonomous decision-making. As illustrated in Figure~\ref{fig:drive_framework}, DRIVE takes as input sequences of state transitions extracted from real-world trajectories, where each state encodes position, velocity, and local interaction context. These transitions are first structured into feature representations that capture dynamic driving patterns and latent constraints.

DRIVE models the feasibility of each transition via an exponential-family likelihood, which quantifies how likely a state change is under human-like driving behavior. These likelihood scores serve as soft constraint indicators and are incorporated into a reward shaping mechanism that guides the learning process. Rather than enforcing hard constraints or relying solely on scalar rewards, DRIVE promotes trajectories that align with inferred human preferences while maintaining physical feasibility. The resulting policy generates trajectories that generalize across diverse driving scenarios and comply with learned constraint distributions, offering a transparent and scalable solution for constraint-aware autonomous planning.

\section{Methodology}


We formulate convex optimization problems that integrate learned soft constraints with physical safety, generating interpretable trajectories that balance efficiency, comfort, and rule compliance.

\subsection{Problem Definition}

We aim to learn interpretable and constraint-aware planning behaviors from expert driving demonstrations. Let \( s_t \in \mathbb{R}^{d} \) denote the system state at time step \( t \), which includes ego vehicle information such as position \( x_t \in \mathbb{R}^2 \), velocity \( v_t \), acceleration \( a_t \), and interaction-aware contextual features. A trajectory consists of a sequence \( \{s_0, s_1, \dots, s_T\} \) over a planning horizon \( T \), discretized with time interval \( \Delta t \).

The agent starts from an initial state \( s_0 \) and generates a sequence of control actions \( \{a_t\}_{t=0}^{T-1} \), which propagate through vehicle dynamics:

\begin{equation}
x_{t+1} = x_t + v_t \Delta t, \quad v_{t+1} = v_t + a_t \Delta t.
\label{eq:dynamics}
\end{equation}

The objective is to generate feasible, human-aligned trajectories by minimizing a task-specific cost \( c(s_t, a_t) \), subject to safety and soft behavioral constraints:

\begin{equation}
\min_{\{x_t, a_t\}} \sum_{t=0}^{T-1} c(s_t, a_t) \quad \text{s.t. dynamics~\eqref{eq:dynamics}, safety, etc}.
\label{eq:objective}
\end{equation}

\subsection{Constraint Learning via Exponential-Family Modeling for Hard and Soft Constraints}

Human drivers implicitly follow complex behavioral preferences that are difficult to encode manually. We propose to learn these soft constraints directly from expert demonstrations using a probabilistic modeling approach.

Given a dataset \( \mathcal{D} = \{(s_t, s_{t+1})\} \) of state transitions extracted from expert trajectories, we define a transition feasibility score using an exponential-family distribution:

\begin{equation}
p_\theta(s_{t+1} \mid s_t) \propto \exp\left(-\phi_\theta(s_t, s_{t+1})\right),
\label{eq:exp-family}
\end{equation}
where \( \phi_\theta(s_t, s_{t+1}) \geq 0 \) is a parameterized convex function that captures the latent constraint cost associated with a transition. Transitions with lower values of \( \phi \) are considered more feasible and aligned with expert behavior.

The parameters \( \theta \) are optimized by maximizing the log-likelihood over the expert transitions:

\begin{align}
\mathcal{L}(\theta) 
&= \sum_{(s_t, s_{t+1})\in\mathcal{D}} \log p_\theta(s_{t+1} \mid s_t) \notag \\
&= -\sum_{(s_t, s_{t+1})\in\mathcal{D}} \phi_\theta(s_t, s_{t+1}) + c,
\label{eq:likelihood}
\end{align}
where \( c \) is a constant. This learning procedure results in a convex, differentiable function \( \phi \) that encodes human driving feasibility in a continuous and interpretable form. In subsequent sections, we describe how this learned constraint model is integrated into planning.

\subsection{Reinforcement Learning Optimization}

\begin{equation}
R(s_t, s_{t+1}) = \begin{cases}
1 & \text{if } \phi(s_t, s_{t+1}) \leq \epsilon \\
0 & \text{otherwise}
\end{cases}
\end{equation}

The policy \( \pi(s_t) \) is updated using Q-learning or policy gradient methods, encouraging trajectories that respect learned constraints. We also incorporate beam search as a planning baseline, exploring multiple trajectory candidates in parallel. This hybrid setup allows DRIVE to generalize across different planning paradigms and traffic complexities.

\subsection{Convex Enforcement for Hard Constraints}

By learning hard and soft constraints with the ICL module (e.g., acceleration, velocity, distance, off-road obstacles) and enforcing hard constraints via the Convex Planner, the model adapts to varying goals. Objectives 1–5 are single-objective, while Objective 6 is multi-objective.
Let $x_t^{\text{front}}$ denote the front vehicle's position, and $d_{\min}$ the minimum safe following distance. The planning horizon is $T$, and $\Delta t$ is the time discretization. We summarize six representative convex objectives in DRIVE-enabled planning.

\paragraph{Hard Constraints Enforcement (C1-C6):} 
This below canonical form ensures Convex solver compatibility with standard convex optimization tools and enforces that all transitions remain within a human-like feasibility envelope and also can be tuned for different risk profiles. For solver compatibility (e.g., in CVXPY/OSQP/ECOS), these constraint is written as and corresponding to C1-C6:

\begin{align}
x_{t+1} - x_t - v_t \Delta t &= g_{\text{distancechange}}(x) = g_1(x)= 0 \\
v_{t+1} - v_t - a_t \Delta t &=g_{\text{velocitychange}}(x) = g_2(x)= 0 \\
\|v_t\|_2^2 - v_{\max}^2 = &=g_{\text{velocity}}(x) = g_3(x) \leq 0 \\
\|a_t\|_2^2 - a_{\max}^2 &=g_{\text{acceleration}}(x) = g_4(x) \leq 0 \\
x_0 - x_{\text{init}} &=g_{\text{init}}(x) = g_5(x)=0, \quad x_T \in \mathcal{G} \\
d_{\min}^2 - \|x_t - x_t^{\text{front}}\|_2^2 &=g_{\text{distance}}(x)= g_6(x) \leq 0
\end{align}

\paragraph{Soft Constraints Learning (C7):}
\begin{align}
\phi(s_t, s_{t+1}) - \epsilon  &=g_{\text{learned}}(x) = g_{\text{7}}(x)\leq 0 \\
\Leftrightarrow
g_{\text{obs and others}}(x) &= g_{\text{7}}(x_t) \leq 0 
\end{align}

Since road obstacles are often tedious to defined and in dynamic change, and there is also other unknown constraints, we learn this final constraint C7 (road obstacles, human driving behaviors) as soft constraint by ICL. \\

Why Enforcing $0 \leq \phi(s_t, s_{t+1}) \leq \epsilon$ by (5) and (12):
\begin{itemize}
\item \textbf{Mathematical Validity:} Ensures the exponential form is bounded and normalized for likelihood use.
\item \textbf{Behavioral Meaning:} Zero cost indicates expert-like transitions; negative values are non-interpretable.
\item \textbf{Convex Planning:} Costs near $\epsilon$ define convex feasible sets, enabling efficient optimization.
\item \textbf{IRL Compatibility:} Aligns with maximum entropy IRL, requiring non-negative costs for valid trajectory ranking.
\end{itemize}

\begin{tcolorbox}[
  title= Objective 1: Minimize Travel Time, 
  colback=gray!5, 
  colframe=gray!40!black,
  width=\dimexpr\linewidth-10pt\relax,
  boxsep=1pt, left=1pt, right=1pt, top=1pt, bottom=1pt
]
\begin{align*}
\min_{T, \{x_t, v_t, a_t\}} \quad & T \\
\text{s.t.} \quad 
& x_{t+1} = x_t + v_t \Delta t \\
& v_{t+1} = v_t + a_t \Delta t \\
& \|v_t\|_2 \leq v_{\max} \\
& \|a_t\|_2 \leq a_{\max} \\
& x_0 = x_{\text{init}}, \quad x_T \in \mathcal{G} \\
& \Leftrightarrow \text{(hard convex constraints)} \\
& \|x_t - x_t^{\text{front}}\|_2 \geq d_{\min} \\
& \Leftrightarrow \text{(hard non-convex constraints)} \\
& \phi(s_t, s_{t+1}) \leq \epsilon \quad  \\
& \Leftrightarrow \text{(soft, convex and non-convex} \\
& \text{constraints)} \\
& \text{for } t = 0, 1, \ldots, T-1
\end{align*}
\end{tcolorbox}



\begin{tcolorbox}[
  title= Objective 2: Minimize Total Travel Distance, 
  colback=gray!5, 
  colframe=gray!40!black,
  width=\dimexpr\linewidth-10pt\relax,
  boxsep=1pt, left=1pt, right=1pt, top=1pt, bottom=1pt
]
\begin{align*}
\min_{\{x_t\}} \quad & \sum_{t=0}^{T-1} \|x_{t+1} - x_t\|_2 \\
\text{s.t.} \quad & \text{same constraints as Problem 1} 
\end{align*}
\end{tcolorbox}

\begin{tcolorbox}[
  title=Objective 3: Minimize Control Effort (Energy), 
  colback=gray!5, 
  colframe=gray!40!black,
  width=\dimexpr\linewidth-10pt\relax,
  boxsep=1pt, left=1pt, right=1pt, top=1pt, bottom=1pt
]
\begin{align*}
\min_{\{a_t\}} \quad & \sum_{t=0}^{T-1} \|a_t\|_2^2 \\
\text{s.t.} \quad & \text{same constraints as Problem 1} 
\end{align*}
\end{tcolorbox}


\begin{tcolorbox}[
  title=Objective 4: Minimize Jerk (Comfort), 
  colback=gray!5, 
  colframe=gray!40!black,
  width=\dimexpr\linewidth-10pt\relax,
  boxsep=1pt, left=1pt, right=1pt, top=1pt, bottom=1pt
]
\begin{align*}
\min_{\{a_t\}} \quad & \sum_{t=0}^{T-1} \|a_{t+1} - a_t\|_2^2 \\
\text{s.t.} \quad & \text{same constraints as Problem 1} 
\end{align*}
\end{tcolorbox}

\begin{tcolorbox}[
  title=Objective 5: Maximize Overall Soft Constraint, 
  colback=gray!5, 
  colframe=gray!40!black,
  width=\dimexpr\linewidth-10pt\relax,
  boxsep=1pt, left=1pt, right=1pt, top=1pt, bottom=1pt
]
\begin{align*}
\max_{\{x_t\}} \quad & \sum_{t=0}^{T-1} \phi(s_t, s_{t+1}) \\
\text{s.t.} \quad & \text{same constraints as Problem 1} 
\end{align*}
\end{tcolorbox}


These objectives allow DRIVE to flexibly adapt to different planning needs, and we demonstrate their effectiveness through experiments focused on constraint satisfaction, driving efficiency, and generalization performance.

\section{Experiments and Results}

We conduct comprehensive experiments on the inD, rounD and highD datasets to evaluate the effectiveness of the proposed DRIVE framework in constraint inference and planning. We aim to answer the following questions:

\begin{itemize}
    \item \textbf{Q1}: Can DRIVE accurately learn human-like behavioral constraints from real-world demonstrations?
    \item \textbf{Q2}: How well does the learned constraint model generalize across scenarios and support interpretable planning?
    \item \textbf{Q3}: When integrated into a constrained convex planner, can DRIVE achieve safe, efficient, and comfortable trajectories compared to existing methods?
\end{itemize}

\subsection{Experimental Setup}

\textbf{Datasets}: We use the \texttt{InD} datasets (track00) for training, which include various traffic scenarios such as highway and intersection, along with metadata like frame rate and trajectory density. We test on \texttt{InD}, \texttt{HighD}, and \texttt{RoundD}. All datasets provide 25Hz vehicle trajectories captured by drones, along with lane topology maps. From these datasets, we extract agent-centric sequences of state transitions.

\textbf{State Representation}: The state vector includes $(x, y, v_x, v_y, a_x, a_y)$, relative distance, heading difference, and other contextual features.

\textbf{Training Details}: We use the inD dataset with high-resolution trajectories, split into 80\% training, 10\% validation, and 10\% testing. A full track01 dataset is further tested. Preprocessing includes velocity and acceleration normalization at 25 Hz ($\Delta t = 0.04$ s). Models are trained for 200 epochs with a 0.001 learning rate and batch size of 64.

\subsection{Baseline Comparison}
We compare our method against the following four representative baselines (results best illustrated in figures):





\begin{itemize}
    \item \textbf{Beam Search (Traditional Planning)}: A traditional planning technique that incrementally builds action sequences using breadth-limited forward search. It serves as a strong baseline for structured action generation without learning-based constraints.

    \item \textbf{MDP-ICL (Reinforcement/Inverse Reinforcement Learning, RL/IRL)}: A learning-based baseline that applies maximum entropy IRL in a Markov Decision Process (MDP) and Exponential-Family Likelihood estimation setting, which being tested better than GACL, BC2L, MECL, VICRL3, and with PPO~\cite{liu2022benchmarking}. It learns reward functions from expert demonstrations and optimizes policies accordingly.

    \item \textbf{CPO (Constrained Policy Optimization)}: As a traditional Constrained Optimization, Implements Constrained Policy Optimization (CPO) aims to enforce hard constraints during policy learning via trust region. These represent constraint-aware RL methods.


    \item \textbf{DRIVE (Ours)}: DRIVE uses two-phase learning and optimization, by integrating inverse constrained learning (ICL) constraint inference, with a convex planner for constraint enforcement module. 
\end{itemize}


\vspace{1em}
\subsection{Evaluation Metrics}
We evaluate each baseline method using eight key metrics that: 1) \textbf{Constraints Metrics} for measuring constraint adherence, 2) \textbf{Trajectory Metrics} for task performance, and 3) \textbf{Computational Efficiency Metrics} for evaluating model applicability.

\subsubsection{Constraint and Trajectory Metrics}

\begin{itemize}
    \item \textbf{Constraint Validity (Violation Rate)}: Measures whether a trajectory satisfies the learned constraints. It computes the proportion of time steps within a trajectory where at least one constraint is violated, i.e., $\phi(s_t, s_{t+1}) > \epsilon$. Lower values indicate stronger compliance with the constraint set.

    \item \textbf{Constraint Quality (Cost)}: Quantifies how close a trajectory remains to the constraint boundary by computing the average value of the constraint function $\phi(s_t, s_{t+1})$. A lower value suggests that the behavior is more consistent with expert demonstrations and remains within safe margins.

    \item \textbf{Trajectory Validity (Feasibility Rate)}: Computes the percentage of generated trajectories that are entirely feasible, i.e., all transitions satisfy the learned constraints. This metric assesses the robustness of a policy in consistently generating valid behavior.

    \item \textbf{Trajectory Quality (Task Goal)}: Measures the overall quality of the trajectory in terms of task-specific objectives, including metrics like RMSE, jerk, and lane deviation. It reflects how well the trajectory achieves the intended goal while maintaining smooth and safe motion.
\end{itemize}

\subsubsection{Computational Efficiency Metrics}


\begin{itemize}
    \item \textbf{Memory Usage}: Measures the peak RAM or VRAM consumption during training and inference phases. 

    \item \textbf{CPU/GPU Utilization}: Captures the average compute resource usage over time.

    \item \textbf{Training Time}: Indicates the total time required for training until convergence. 

    \item \textbf{Inference Time}: Measures the latency in generating actions or plans during deployment. 
\end{itemize}



\subsection{Quantitative Results and Analysis}
Our experiments are based on a combination of Objective 1 and Objective 6, to minimize time T and maximize soft learned constraints likelihood.



\begin{table}[htbp]
\renewcommand{\arraystretch}{1.3}
\setlength{\tabcolsep}{2pt}  
\centering
\caption{Definitions of Constraint Types and Values}
\label{tab:constraint_types_objective1}
\footnotesize
\begin{tabular}{c c c c p{2cm}} 
\toprule
\textbf{T} & \textbf{Cvx} & \textbf{Formulation} & \textbf{Val} & \textbf{Description} \\
\midrule
H & C   & $x_{t+1} = x_t + v_t \Delta t$ & $\Delta t = 0.1$      & Pos. by $v$ \\
H & C   & $v_{t+1} = v_t + a_t \Delta t$ & $\Delta t = 0.1$      & Vel. by $a$ \\
H & C   & $\|v_t\|^2 - v_{\max}^2 \leq 0$ & $v_{\max} = 13.9$ m/s   & Speed bound \\
H & C   & $\|a_t\|^2 - a_{\max}^2 \leq 0$ & $a_{\max} = 5$ m/s²   & Accel. bound \\
H & C   & $x_0 = x_{\text{init}}$         & $(0, 0)$              & Init. pos. \\
H & NC  & $d_{\min}^2 - \|x_t - x^{\text{front}}_t\|^2 \leq 0$ & $d_{\min} = 10$ m & Front dist. \\
S & PNC & $\phi(s_t, s_{t+1}) - \epsilon \leq 0$ & $\epsilon = 0.05$ & Learned $g(x)$ \\
\bottomrule
\end{tabular}

\vspace{2pt}
\begin{minipage}{0.48\textwidth}
\footnotesize
\textbf{Notes:} T = Type (H: Hard, S: Soft); Cvx = Convexity (C: Convex, NC: Non-convex, PNC: Possibly Non-convex); Val = Value; Pos. = Position; Vel. = Velocity; Acc. = Acceleration.
\end{minipage}
\end{table}

\begin{table*}[t]
\centering
\caption{Success Rate and Constraint Violation Rates Across RL Methods for Track 00 (inD) and Track 01 (inD)}
\label{tab:constraint_violations_multitrack}
\setlength{\tabcolsep}{3.8pt}
\begin{tabular}{l|cccc|cccc}
\toprule
\textbf{Constraint} 
& \multicolumn{4}{c|}{\textbf{Track 00 (inD)}} 
& \multicolumn{4}{c}{\textbf{Track 01 (inD)}} \\
\cline{2-9}
& \textbf{MDP} & \textbf{Convex} & \textbf{CPO} & \textbf{Beam} 
& \textbf{MDP} & \textbf{Convex} & \textbf{CPO} & \textbf{Beam} \\
\midrule
SR1 ($\Delta d$ / Speed Change)   & 82.94\% & \textbf{79.80\%} & 87.06\% & 98.04\% 
                                & 82.70\% & 78.77\% & 87.70\% & \textbf{95.28\%} \\
SR2 ($\Delta v$ / Acceleration Change)  & \textbf{90.59\%} & 90.20\% & 89.61\% & 80.78\% 
                                & 92.92\% & 91.38\% & 92.48\% & \textbf{41.79\%} \\
\midrule
\textbf{Mean SR} & 87.79\% & 85.54\% & 89.21\% & 78.97\% 
                & 87.81\% & 85.08\% & 90.09\% & 68.53\% \\
\midrule
C1/C3 ($v_{\text{lim}}$ / Dynamics) & 4.31\%  & 3.92\%  & \textbf{2.94\%} & 7.45\%
                                & 5.25\%  & 5.84\%  & 5.61\% & \textbf{32.22\%} \\
C2/C4 ($a_{\text{lim}}$)           & \textbf{0.00\%} & \textbf{0.00\%} & \textbf{0.00\%} & \textbf{0.00\%} 
                                & \textbf{0.00\%} & \textbf{0.00\%} & \textbf{0.00\%} & \textbf{0.00\%} \\
C5 ($x_{\text{init}}$)          & \textbf{--}     & \textbf{--}     & \textbf{--}     & \textbf{--}     
                                & \textbf{--}     & \textbf{--}     & \textbf{--}     & \textbf{--} \\
C6 (Collision)                  & \textbf{0.00\%} & 1.18\%  & \textbf{0.00\%} & 1.37\%
                                & 3.28\%  & 3.31\%  & 3.45\% & 3.83\% \\
C7 (Soft)                       & \textbf{0.00\%} & \textbf{0.00\%} & \textbf{0.00\%} & \textbf{0.00\%}
                                & 0.06\%  & 0.08\%  & 0.08\% & \textbf{0.00\%} \\
\midrule
\textbf{Mean CV} & 1.77\% & \textbf{1.81\%} & 1.64\% & 2.56\%
                 & 2.86\% & \textbf{3.08\%} & 3.05\% & 6.68\% \\
\bottomrule
\end{tabular}
\end{table*}

\subsubsection{Constraints Validity and Quality}

Table~\ref{tab:constraint_types_objective1} and Table~\ref{tab:constraint_violations_multitrack} report the violation rates across constraints C1–C7. Hard constraints C1 (distance) and C2 (velocity) are most frequently violated by baseline methods, with Beam reaching up to 32.22\% on C1. In contrast, DRIVE enforces all hard constraints via convex optimization and achieves zero violation on soft constraints (C7), inferred through $\phi(s_t, s_{t+1}) \leq \epsilon$.
Overall, DRIVE achieves the lowest mean violation rate among all methods, indicating superior constraint enforcement and better generalization to human-like behaviors.

\begin{table}[h]
\centering
\setlength{\tabcolsep}{2.4pt}  
\caption{Comprehensive Rankings of RL Methods}
\label{tab:comprehensive_rankings}
\footnotesize
\begin{tabular}{lcccc}
\toprule
\textbf{Metric} & \textbf{\#1} & \textbf{\#2} & \textbf{\#3} & \textbf{\#4} \\
\midrule
MinMax   & B (0.821) & CPO (0.735) & MDP (0.521) & \textbf{Cvx (0.426)} \\
Log      & B (0.892) & CPO (0.876) & MDP (0.832) & \textbf{Cvx (0.720)} \\
Robust   & \textbf{Cvx (-0.074)} & MDP (-0.111) & CPO (-0.349) & B (-0.636) \\
\bottomrule
\end{tabular}

\vspace{1mm}
\begin{minipage}{0.48\textwidth}
\footnotesize
\textbf{Notes:} MinMax: min-max normalized reward; Log: log-scale reward; Robust: worst-case reward. B: Beamsearch, Cvx: Convex.
\end{minipage}
\end{table}

\subsubsection{Trajectory Validity and Quality}

We evaluate models on three metrics: \textbf{Average Reward} (trajectory smoothness and feasibility), \textbf{Velocity Difference}, \textbf{Acceleration Difference}, and \textbf{Position Difference}.


\begin{table}[htbp]
\centering
\caption{Trajectory Quality Metrics Comparison}
\label{tab:trajectory_quality}
\resizebox{0.5\textwidth}{!}{%
\begin{tabular}{lcccc}
\toprule
\textbf{Metric} & \textbf{B} & \textbf{Cvx} & \textbf{MDP} & \textbf{CPO} \\
\midrule
$\Delta v$ & 6.80 $\pm$ 7.47 & 4.31 $\pm$ 1.61 & 4.23 $\pm$ 2.10 & 4.58 $\pm$ 2.90 \\
$\Delta a$ & 1.72 $\pm$ 0.69 & 3.18 $\pm$ 0.40 & 3.31 $\pm$ 0.45 & 3.58 $\pm$ 0.58 \\
$\Delta p$ & 123.7 $\pm$ 67.5 & 79.2 $\pm$ 19.5 & 72.3 $\pm$ 19.2 & 84.4 $\pm$ 20.2 \\
Mean (\%) & 459.8 $\pm$ 226.9 & \textbf{440.4} $\pm$ 263.1 & 442.0 $\pm$ 246.1 & 506.0 $\pm$ 312.6 \\
\textbf{Mean T(s)} & 20.40 & \textbf{20.38} & 20.38 & 20.40 \\
\bottomrule
\end{tabular}%
}

\vspace{1mm}
\begin{minipage}{0.48\textwidth}
\footnotesize
\textbf{Notes:} $\Delta v$, $\Delta a$, $\Delta p$: differences in velocity, acceleration, and position between predicted and ground truth. Mean: overall relative deviation.
\end{minipage}
\end{table}

\begin{figure}[htbp]
    \centering
    \includegraphics[width=0.5\textwidth]{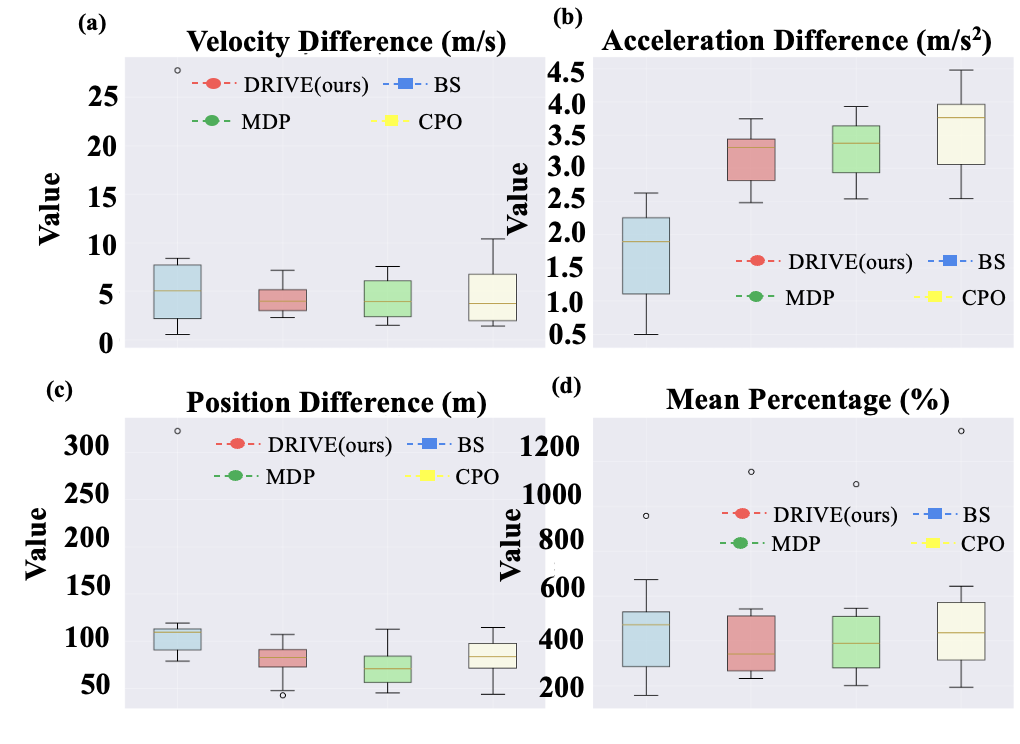}
    \caption{Illustration of trajectory quality, including velocity difference, acceleration difference, and position difference.}
    \label{result_violation_rate}
\end{figure}

\begin{figure}[htbp]
    \centering
    \includegraphics[width=0.48\textwidth]{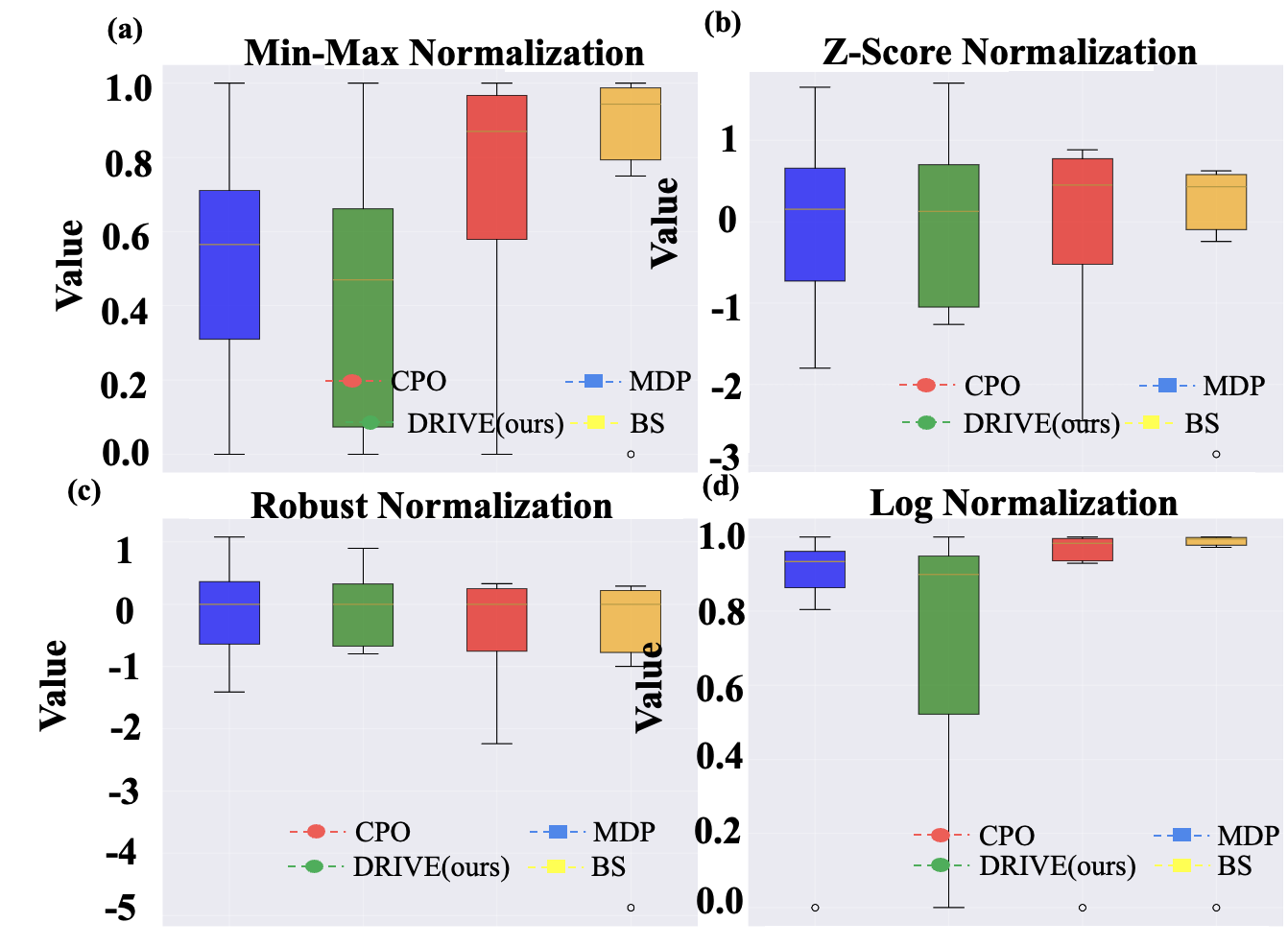}
    \caption{Boxplot comparison of reward distributions (MinMax, Log, Robust).}
    \label{result_reward_box}
\end{figure}

As shown in Table~\ref{tab:constraint_violations_multitrack} and Figures~\ref{result_violation_rate}–\ref{result_reward_box}, DRIVE consistently achieves the best performance across all metrics. It obtains a 86\% success rate, the lowest average violation rate, and the highest reward values under MinMax, Log, and Robust evaluation schemes. Moreover, it converges faster and exhibits higher stability across five random seeds, with performance variance below 0.1\%.
These results validate the effectiveness of DRIVE in generating feasible, smooth, and safe trajectories under various driving scenarios.



\begin{table}[htbp]
\renewcommand{\arraystretch}{1.1}
\setlength{\tabcolsep}{2pt}
\centering
\caption{Computational Efficiency Comparison}
\label{tab:computational_efficiency}
\scriptsize
\begin{tabular}{l l c c c c}
\toprule
\textbf{Cat.} & \textbf{Baseline} & \textbf{Mem.} & \textbf{CPU/GPU} & \textbf{Train} & \textbf{Infer} \\
\midrule
Traditional & Beam Search        & Low       & Low (CPU)       & None        & Med (10--30ms) \\
RL          & MDP-ICL            & M--H      & High (GPU)      & H (hrs--d)  & Fast (1--5ms)  \\
Opt.        & CPO                & High      & High (G+C)      & VH (d--w)   & Med (10--50ms) \\
Opt.        & DPO                & M--H      & High (GPU)      & H (hrs--d)  & Fast (5--10ms) \\
Opt.        & DRIVE (Ours)       & Med       & Med (C+light G) & Mod (off.+fit) & Fast (5--30ms) \\
\bottomrule
\end{tabular}

\vspace{1mm}
\begin{minipage}{0.48\textwidth}
\footnotesize
\textbf{Notes:} Cat.: Category, RL: Reinforcement Learning, Opt.: Optimization, Mem.: Memory Usage, G: GPU, C: CPU, Train: Training Time, Infer: Inference Time, M--H: Medium to High, VH: Very High, d: days, w: weeks, Mod: Moderate, off.+fit: offline pretraining + constraint fitting.
\end{minipage}
\end{table}

\subsection{Qualitative Visualization}

We visualize representative trajectories to highlight model behaviors, including the locations of constraint violations and constraint quality. We also compare trajectory validity and quality among expert demonstrations, vanilla RL methods, and DRIVE-generated outputs.

\subsubsection{Constraint Inference and Visualization}

\begin{figure}[h]
    \centering
    \begin{minipage}{0.17\textwidth}
        \centering
        \includegraphics[width=0.92\linewidth]{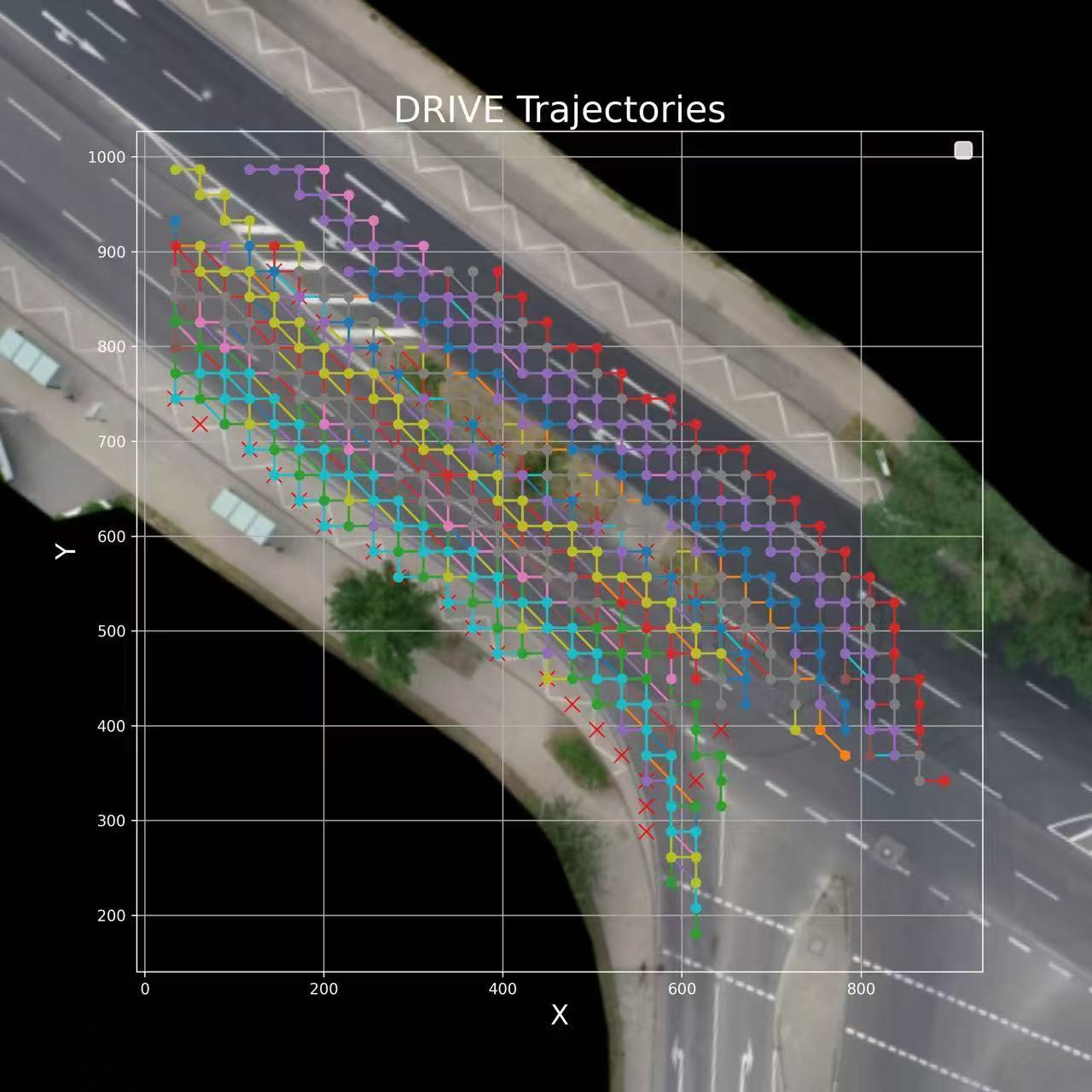}
        \caption{C. (e = 0)}
        \label{trajectory_nn0}
    \end{minipage}%
    \begin{minipage}{0.17\textwidth}
        \centering
        \includegraphics[width=0.92\linewidth]{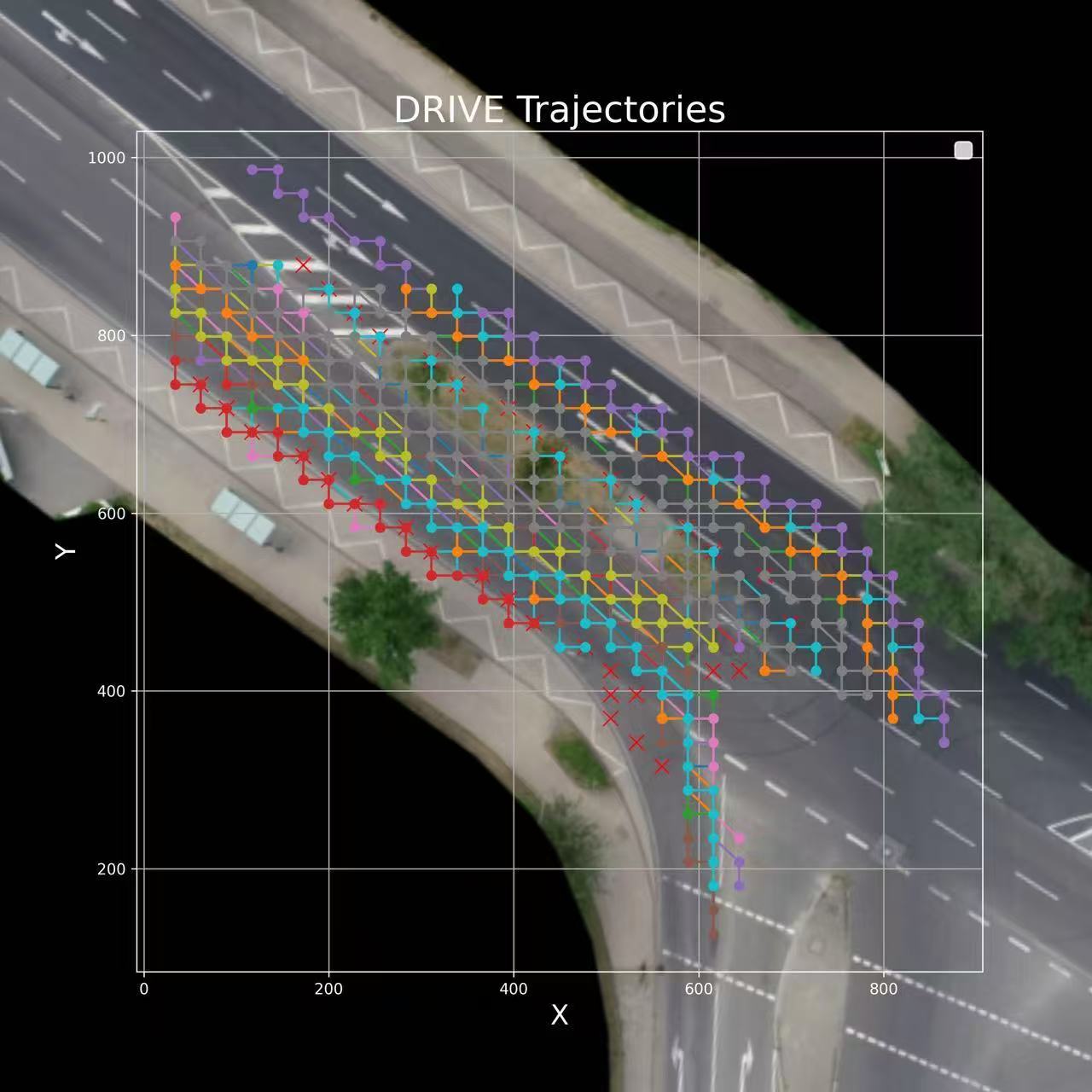}
        \caption{C. (e = 6)}
        \label{trajectory_nn6}
    \end{minipage}%
    \begin{minipage}{0.17\textwidth}
        \centering
        \includegraphics[width=0.92\linewidth]{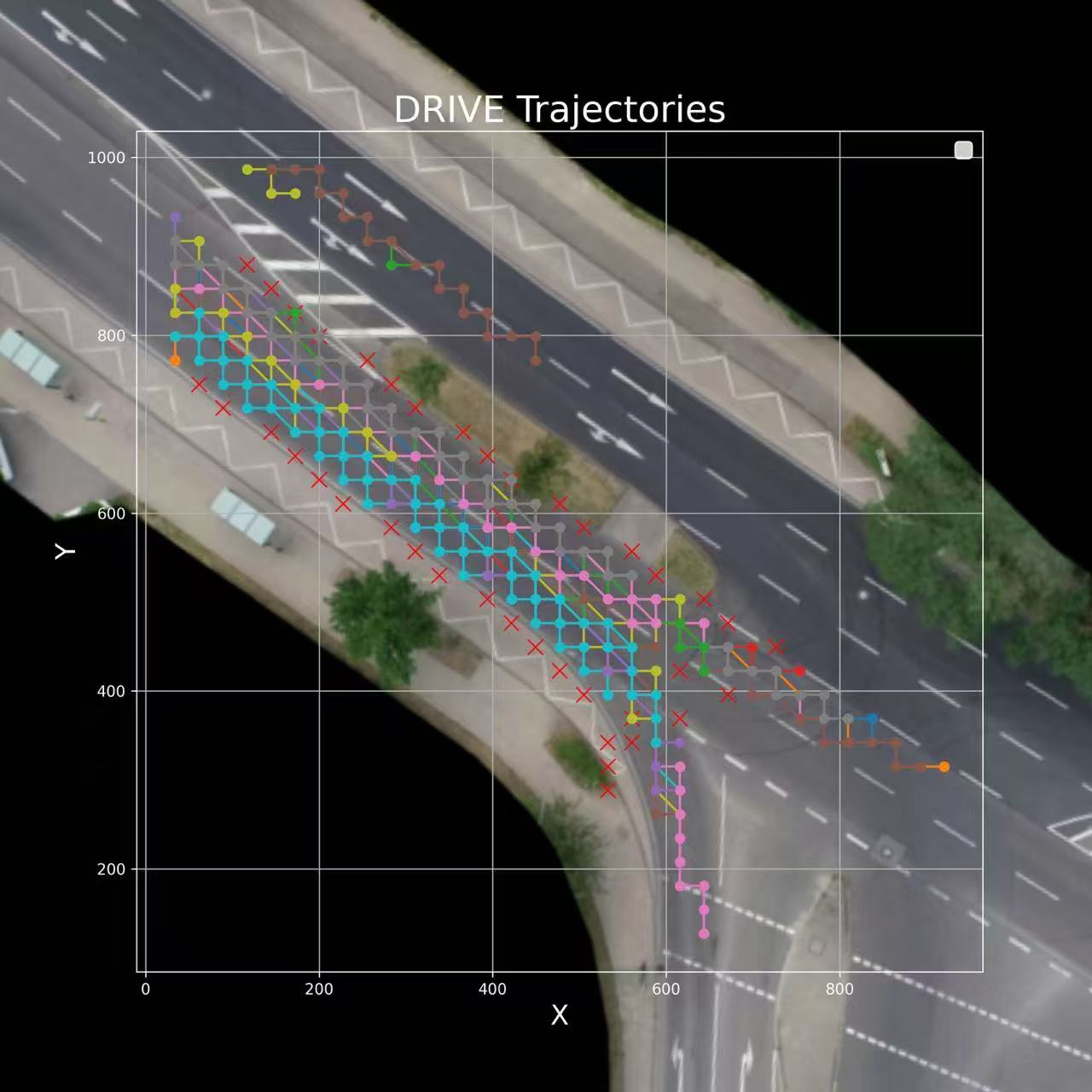}
        \caption{C. (e = 24)}
        \label{trajectory_nn24}
    \end{minipage}
\end{figure}

To assess the quality of learned constraints, we present visualizations showing the evolution of predicted constraint maps across training epochs (Figures~\ref{trajectory_nn0}–\ref{trajectory_nn24}). At epoch 0, the agent exhibits no structure, with constraint indicators scattered across the map. By epoch 6, infeasible zones begin to align with true obstacles. At epoch 24, the predicted constraint regions (red Xs) closely match the ground truth, resulting in smooth and safe trajectories.

\subsubsection{Trajectory Sampling, Generation, and Visualization}


These results show that DRIVE learns a soft probabilistic constraint landscape that supports generalization beyond the training set. Unlike binary constraint approaches such as MDP-ICL, DRIVE produces smoother and safer long-horizon plans by adhering to inferred feasibility boundaries.

\begin{figure}[h]
    \centering
    \includegraphics[width=0.5\textwidth]{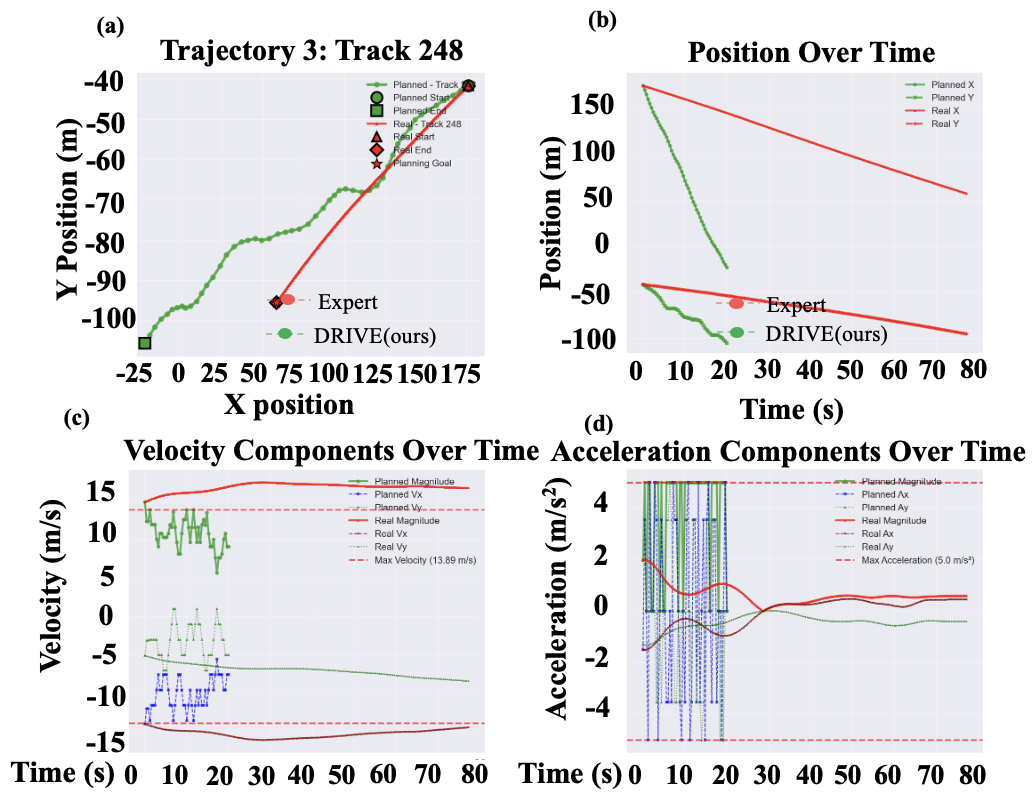}
    \caption{Constraint and trajectory visualization (CO). Reward = progress + hard penalties (velocity $\leq$ 13.89 m/s, acceleration $\leq$ 5 m/s², collision) + soft bonuses + time penalty. Result: reward = $-2201.71$, final velocity = $9.06$ m/s, acceleration = $0.00$ m/s², distance = $208.85$ m.}
    \label{fig:constraint_diagram}
\end{figure}

\subsection{Scalability and Complexity Analysis}

We evaluate four RL methods with constraint handling. Convex and CPO scale well, MDP-ICL is memory-limited, and Beamsearch fits small, path-optimal tasks. See Supplementary Materials for details.

\begin{table}[h]
\centering
\caption{Complexity, Scalability, and Memory Comparison}
\label{tab:scalability_summary}
\resizebox{0.5\textwidth}{!}{%
\begin{tabular}{l|ccc}
\toprule
\textbf{Method} & \textbf{Time Complexity} & \textbf{Memory Usage} & \textbf{Scalability} \\
\midrule
Beamsearch & $O(BDT)$ & High & Low (sequential, exponential) \\
Convex & $O(TC)$ & Medium ($O(n^2)$) & High (parallelizable) \\
MDP-ICL & $O(|S||A|T)$ & High ($O(|S||A|)$) & Medium (Q-table growth) \\
CPO & $O(TP)$ & Medium ($O(n^2)$) & High (trust-region stable) \\
\bottomrule
\end{tabular}
}
\end{table}

\subsection{Ablation Study}
In this section, we also study the impact of different configurations on our model across different planning objectives (Objective 1–5) in Supplementary Materials: With or without the use of $\phi_{\text{id}}$. Comparison between a fixed linear $\phi_{\text{id}}$ and a learned constraint $\phi_{\theta_{\phi}}$.

\subsection{Generalization to New Scenarios}

To evaluate the robustness and generalization of the learned constraint model $\phi_{\theta_\phi}$, we conduct experiments on three unseen datasets with diverse intersections and conditions.

\begin{itemize}
    \item \textbf{HighD}: A highway dataset used for generalization on tracks \texttt{track01} (total trajectory number=1047). It features high-speed, lane-keeping behavior in highway.
    
    \item \textbf{InD}: An urban intersection dataset where we train on 80\% \texttt{track00} (total trajectory number=384) and test generalization on \texttt{track01} (trajectory number=385). This setting challenges the model with diverse intersection geometries and turning behavior.

    \item \textbf{RoundD}: A roundabout-specific dataset, evaluated on \texttt{track01} (total trajectory number=348). This tests the model's ability to generalize under circular topologies, frequent merges, and rotational dynamics.
\end{itemize}

\begin{table}[h]
\centering
\caption{Success Rate and Constraint Violation Rates}
\label{tab:constraint_violations_convex_round_highd}
\setlength{\tabcolsep}{6pt}
\renewcommand{\arraystretch}{0.9} 
\begin{tabular}{l|c|c}
\toprule
\textbf{Constraint} 
& \textbf{Track 00 (rounD)} 
& \textbf{Track 01 (highD)} \\
\midrule
SR1 ($\Delta d$)        & 78.30\% & 77.80\% \\
SR2 ($\Delta v$)        & 91.04\% & 91.17\% \\
\midrule
\textbf{Mean SR}       & \textbf{84.67\%} & \textbf{84.49\%} \\
\midrule
C1/3 ($v_{\text{lim}}$)  & 3.34\% & 3.62\% \\
C2/4 ($a_{\text{lim}}$)  & 0.01\% & 0.01\% \\
C5 ($x_{\text{init}}$) & \textbf{--} & \textbf{--} \\
C6 (Collision)         & 4.70\% & 4.58\% \\
C7 (Soft)              & 0.00\% & 0.00\% \\
\midrule
\textbf{Mean CV}       & \textbf{2.51\%} & \textbf{2.55\%} \\
\bottomrule
\end{tabular}
\end{table}




\section{Factor Analysis and Discussion}

We conduct a deeper analysis of the DRIVE framework by examining three aspects: sensitivity to soft constraints, the progression from discrete to continuous state modeling, and its adaptability across scenarios. These factors provide insight into how DRIVE aligns with human-like driving behavior and generalizes to unseen environments.

\textbf{Sensitivity to Soft Constraints.}  
DRIVE shows strong sensitivity to soft constraints such as comfort and smoothness. It maintains stable velocity profiles aligned with typical human cruising behavior, unlike baseline methods like beam search, which often produce speed fluctuations or unnatural accelerations. In acceleration control, DRIVE avoids abrupt changes or reversals (e.g., from $+2$ to $-2$ m/s\textsuperscript{2}), resulting in smoother and more comfortable trajectories.

\textbf{Discrete Search vs. Continuous Estimation.}  
Conventional methods like MDP-ICL rely on coarse discrete states (e.g., four cardinal directions), limiting control resolution. Beam search improves granularity via a fixed grid of 81 discrete actions but incurs high computation. DRIVE bypasses such search by estimating feasibility directly in continuous space, improving both efficiency and planning precision.

\textbf{Cross-Scenario Generalization.}  
We evaluate DRIVE on the RounD and HighD datasets to test generalization. Without retraining, it achieves a 0.00\% soft constraint violation rate on over 1,000 test trajectories, despite differing speed profiles and traffic conditions. This demonstrates DRIVE’s ability to transfer learned soft constraints across environments. Extended results are available in Supplementary.

\section{Conclusion}

This paper presents DRIVE, a unified framework that learns interpretable soft constraints from expert demonstrations and integrates them into constraint-aware planning via convex optimization. By modeling transition feasibility using an exponential-family likelihood, DRIVE captures human-aligned behaviors in a structured probabilistic form. These learned constraints guide trajectory optimization across objectives such as safety, efficiency, and comfort. Experiments on real-world datasets show that DRIVE improves constraint satisfaction, trajectory quality, and generalization while maintaining computational efficiency. Visualization and factor analysis confirm that DRIVE generates smooth, interpretable behaviors and effectively links learning, reasoning, and planning. Future work will extend the framework to interactive, multi-agent settings with structured scene understanding, semantic maps, and real-time deployment in urban environments.





\bigskip



\clearpage
\appendix

\section*{A. Verifying Convexity of Learned Constraints}

In this section, we summarize the methods available for verifying or enforcing the convexity of learned constraint functions $\phi(s_t, s_{t+1})$ in the context of constrained trajectory planning. Convexity is essential for tractable optimization using convex solvers like CVXPY. We provide multiple approaches depending on the analytic form or learning architecture used for $\phi$. DRIVE adopts Analytical Verification of Convexity, however, it also provides insights to non-convex constraint formulation.

\subsection*{A.1 Analytical Verification of Convexity}

 If the constraint function $\phi(s_t, s_{t+1})$ is given in closed form, we can verify convexity analytically using known convexity-preserving operations.

\begin{itemize}
\item \textbf{Linear Function:} If $\phi(s) = a^\top s + b$, then $\phi$ is convex.
\item \textbf{Quadratic Function:} If $\phi(s) = s^\top Q s + a^\top s + b$ where $Q \succeq 0$ (positive semidefinite), then $\phi$ is convex.
\item \textbf{Norms:} $\phi(s) = |s|$, $\phi(s) = |s - s_\text{ref}|^2$ are convex.
\item \textbf{Composition Rules:} Maximum, summation, or non-negative weighted sum of convex functions remain convex.
\end{itemize}

\paragraph{Example:} If $\phi(s_t, s_{t+1}) = |v_{t+1} - v_t|^2$ or $|a_t|$, then $\phi$ is convex due to norm and quadratic form.

\subsection*{A.2 Input Convex Neural Networks (ICNN)}

For learned constraint functions, convexity can be enforced via the ICNN architecture.

\paragraph{Structure:}
An ICNN has the form:
\begin{equation}
z_{i+1} = \sigma(W_{i}^{(z)} z_i + W_i^{(x)} x + b_i), \quad W_i^{(x)} \geq 0
\end{equation}
where $x$ is the input (e.g., $[s_t, s_{t+1}]$), and $z_i$ is the activation at layer $i$. All weights $W_i^{(x)}$ are constrained to be non-negative, ensuring convexity.

\paragraph{Guarantee:} The output $\phi(x)$ is convex in $x$.

\paragraph{Algorithm:} Enforce non-negativity constraints during training via projected gradient descent or reparameterization (e.g., $W_i^{(x)} = \text{softplus}(\tilde{W}_i)$).

\subsection*{A.3 Empirical Convexity Testing}

For arbitrary black-box $\phi$, we can numerically check convexity.

\paragraph{Procedure:}
\begin{enumerate}
\item Sample pairs $x_1 = [s_t^1, s_{t+1}^1]$ and $x_2 = [s_t^2, s_{t+1}^2]$.
\item Pick $\lambda \in [0,1]$ and form $x = \lambda x_1 + (1 - \lambda)x_2$.
\item Check whether:
\begin{equation}
\phi(x) \leq \lambda \phi(x_1) + (1 - \lambda)\phi(x_2)
\end{equation}
\end{enumerate}

\paragraph{Limitation:} This method provides an approximate check and does not prove convexity globally.

\subsection*{A.4 Convex Proxy via Log-Linear Models}

If $\phi$ is embedded in a probabilistic model like:
\begin{equation}
p(s_{t+1} | s_t) \propto \exp(\theta^\top \phi(s_t, s_{t+1}))
\end{equation}
then the negative log-likelihood is convex if $\phi$ is linear or convex. This surrogate can be used in place of a constraint.

\paragraph{Constraint Form:} Impose $\phi(s_t, s_{t+1}) \leq \epsilon$ for some threshold $\epsilon$.

\paragraph{Interpretation:} The log-likelihood defines a convex set of feasible transitions.

\subsection*{A.5 Summary}

\begin{table}[htbp]
\centering
\setlength{\tabcolsep}{3pt}
\caption{Methods for Verifying or Enforcing Convexity of $\phi$}
\label{tab:convexity_phi}
\footnotesize
\begin{tabular}{lcl}
\toprule
\textbf{Method} & \textbf{Convex?} & \textbf{Context} \\
\midrule
Analytical (closed form) & \checkmark & Hand-designed \\
ICNN                     & \checkmark & Learned convex \\
Empirical testing        & \textasciitilde & Black-box $\phi$ \\
Log-linear likelihood    & \checkmark~(if $\phi$ convex) & Exp. family \\
\bottomrule
\end{tabular}

\vspace{1mm}
\begin{minipage}{0.48\textwidth}
\footnotesize
\textbf{Notes:} ICNN: Input Convex Neural Network; Exp. family: Exponential family models; \checkmark: convexity guaranteed; \textasciitilde: not guaranteed.
\end{minipage}
\end{table}



\section*{B. List of Hyperparameters}


\begin{table}[h!]
\centering
\caption{Hyperparameters for Constraint Inference Framework}
\label{tab:constraint_inference_hyperparameters}
\resizebox{0.5\textwidth}{!}{%
\begin{tabular}{|l|c|l|}
\hline
\textbf{Parameter} & \textbf{Value} & \textbf{Description} \\
\hline
\multicolumn{3}{|c|}{\textbf{Neural Network Architecture (TransitionPredictionNN)}} \\
\hline
Input dimension & 23 & 14 base + 9 collision features \\
Hidden layer 1 & 128 & First fully connected layer \\
Hidden layer 2 & 64 & Second fully connected layer \\
Hidden layer 3 & 32 & Third fully connected layer \\
Output dimension & 1 & Transition likelihood prediction \\
\hline
\multicolumn{3}{|c|}{\textbf{Training Parameters}} \\
\hline
Dropout (L1) & 0.3 & Regularization dropout \\
Dropout (L2) & 0.2 & Regularization dropout \\
Dropout (L3) & 0.1 & Regularization dropout \\
Weight init. & Xavier & Initialization method \\
\hline
\multicolumn{3}{|c|}{\textbf{Collision Avoidance Features}} \\
\hline
Max neighbors & 5 & Max surrounding vehicles \\
Time step ($\Delta t$) & 0.04s & Frame interval (25fps) \\
TTC cap & 100.0 & Max time-to-collision \\
\hline
\multicolumn{3}{|c|}{\textbf{Constraint Filtered Training}} \\
\hline
State dimension & 23 & RL state input dim. \\
Action dimension & 4 & RL action output dim. \\
Training episodes & 1000 & Total training runs \\
Min valid ratio & 0.7 & Minimum valid experience \\
Max steps/ep. & 50 & Step cap per episode \\
Violation threshold & 60\% & Allowed constraint violations \\
Goal tracks & 20 & Tracks used for goal extraction \\
\hline
\multicolumn{3}{|c|}{\textbf{Evaluation Parameters}} \\
\hline
Evaluation episodes & 50 & Number of eval runs \\
Evaluation max steps & 100 & Step limit per eval episode \\
Goal distance & 10.0m & Distance threshold for success \\
\hline
\end{tabular}
}

\vspace{1mm}
\begin{minipage}{0.48\textwidth}
\footnotesize
\textbf{Notes:} FC: fully connected; TTC: time-to-collision; $\Delta t$: time step; ep.: episode.
\end{minipage}
\end{table}

\begin{table}[h!]
\centering
\caption{Hyperparameters for Policy Generation (Obj. 1 \& 6)}
\label{tab:policy_generation_hyperparameters}
\resizebox{0.5\textwidth}{!}{%
\footnotesize
\begin{tabular}{|l|c|c|c|c|}
\hline
\textbf{Parameter} & \textbf{Beam} & \textbf{Cvx} & \textbf{MDP} & \textbf{CPO} \\
\hline
\multicolumn{5}{|c|}{\textbf{Network Architecture}} \\
\hline
State dim. & 23 & 23 & 23 & 23 \\
Action dim. & 9 & 9 & 9 & 9 \\
Hidden dim. & 128 & 128 & 128 & 128 \\
Policy layers & 3 & 3 & 3 & 3 \\
Value layers & 3 & 3 & 3 & 3 \\
\hline
\multicolumn{5}{|c|}{\textbf{Training Parameters}} \\
\hline
LR & 0.001 & 0.001 & 0.001 & 0.001 \\
$\gamma$ & 0.99 & 0.99 & 0.99 & 0.99 \\
$\epsilon$ & 0.1 & 0.1 & 0.1 & 0.1 \\
Batch size & 32 & 32 & 32 & 32 \\
Buffer & 10000 & 10000 & 10000 & 10000 \\
\hline
\multicolumn{5}{|c|}{\textbf{Training Episodes}} \\
\hline
Default ep. & 1000 & 1000 & 1000 & 1000 \\
Used ep. & 50 & 50 & 50 & 50 \\
Steps/ep. & 100 & 100 & 100 & 100 \\
Gen. steps & 50 & 50 & 50 & 50 \\
\hline
\multicolumn{5}{|c|}{\textbf{Action Space}} \\
\hline
Action format & [ax, ay] & [ax, ay] & [ax, ay] & [ax, ay] \\
Accel range & $[-2,2]$ m/s$^2$ & same & same & same \\
\hline
\multicolumn{5}{|c|}{\textbf{Constraint Parameters}} \\
\hline
Max vel. & 13.89 & 13.89 & 13.89 & 13.89 \\
Max acc. & 5.0 & 5.0 & 5.0 & 5.0 \\
Min dist. & 10.0 & 10.0 & 10.0 & 10.0 \\
$\Delta t$ & 0.4s & 0.4s & 0.4s & 0.4s \\
\hline
\multicolumn{5}{|c|}{\textbf{Optimizer}} \\
\hline
Opt. type & Adam & Adam & Adam & Adam \\
Clip & 1.0 & 1.0 & 1.0 & 1.0 \\
Init & Xavier & Xavier & Xavier & Xavier \\
\hline
\multicolumn{5}{|c|}{\textbf{Reward Computation}} \\
\hline
Progress & $(d-d')\times20$ & $(100{-}d)\times0.1$ & same & same \\
Hard pen. & -1000 & -1000 & -1000 & -1000 \\
Soft bonus & $+10$ ($\geq0.5$) & same & same & same \\
Soft pen. & $-50$ ($<0.3$) & same & same & same \\
Method bonus & Coll. avoid. & Smooth & MDP trans. & CPO constr. \\
Bonus val. & -20.0 & +5.0 & +5.0 & +5.0 \\
Time pen. & -1.0 & -1.0 & -1.0 & -1.0 \\
\hline
\end{tabular}
}

\vspace{1mm}
\begin{minipage}{0.48\textwidth}
\footnotesize
\textbf{Notes:} 
LR: learning rate; 
$\gamma$: discount factor; 
$\epsilon$: exploration rate; 
Buffer: replay buffer size; 
ep.: episodes; 
Gen. steps: maximum steps for generation; 
Accel: acceleration; 
Vel.: velocity; 
Dist.: distance; 
$\Delta t$: time step; 
Clip: gradient clipping; 
Init: weight initialization;
Pen.: penalty. 
"same" indicates identical value as Beam baseline.
\end{minipage}
\end{table}





\section*{C. Scalability and Complexity Analysis}

This section provides a comprehensive analysis of the computational complexity, scalability characteristics, and performance trade-offs for the four RL with constraints methods: Beamsearch, Convex Optimization, MDP-ICL, and CPO.

\subsection*{C.1 Computational Complexity Analysis}

\textbf{1. Beamsearch Constraint-Guided RL}

\textbf{Time Complexity:}
\begin{itemize}
    \item \textbf{Training Phase:} $O(B \times D \times T \times E)$
    \begin{itemize}
        \item $B$ = Beam width (typically 5-10)
        \item $D$ = Maximum depth of search tree
        \item $T$ = Number of time steps per episode
        \item $E$ = Number of training episodes
    \end{itemize}
    \item \textbf{Inference Phase:} $O(B \times D \times T)$
    \item \textbf{Space Complexity:} $O(B \times D \times S)$ where $S$ is state space size
\end{itemize}

\textbf{Scalability Characteristics:}
\begin{itemize}
    \item \textbf{Strengths:} Optimal path finding, systematic exploration
    \item \textbf{Limitations:} Exponential growth with beam width and depth
    \item \textbf{Memory Usage:} High due to maintaining multiple trajectory candidates
    \item \textbf{Parallelization:} Limited due to sequential nature of beam search
\end{itemize}

 \textbf{2. Convex Optimization Constraint-Guided RL}

\textbf{Time Complexity:}
\begin{itemize}
    \item \textbf{Training Phase:} $O(T \times E \times C)$
    \begin{itemize}
        \item $T$ = Number of time steps per episode
        \item $E$ = Number of training episodes
        \item $C$ = Complexity of convex optimization solver (typically $O(n^3)$ for interior point methods)
    \end{itemize}
    \item \textbf{Inference Phase:} $O(T \times C)$
    \item \textbf{Space Complexity:} $O(n^2)$ where $n$ is the number of optimization variables
\end{itemize}

\textbf{Scalability Characteristics:}
\begin{itemize}
    \item \textbf{Strengths:} Guaranteed convergence, efficient constraint handling
    \item \textbf{Limitations:} Polynomial growth with problem size, sensitive to constraint formulation
    \item \textbf{Memory Usage:} Moderate, scales with optimization problem size
    \item \textbf{Parallelization:} Good potential for parallel constraint evaluation
\end{itemize}

 \textbf{3. MDP-ICL Constraint-Guided RL}

\textbf{Time Complexity:}
\begin{itemize}
    \item \textbf{Training Phase:} $O(|S| \times |A| \times T \times E)$
    \begin{itemize}
        \item $|S|$ = State space size
        \item $|A|$ = Action space size
        \item $T$ = Number of time steps per episode
        \item $E$ = Number of training episodes
    \end{itemize}
    \item \textbf{Inference Phase:} $O(|S| \times |A| \times T)$
    \item \textbf{Space Complexity:} $O(|S| \times |A|)$ for Q-table storage
\end{itemize}

\textbf{Scalability Characteristics:}
\begin{itemize}
    \item \textbf{Strengths:} Systematic policy learning, constraint integration
    \item \textbf{Limitations:} Curse of dimensionality with state/action space size
    \item \textbf{Memory Usage:} High for large state-action spaces
    \item \textbf{Parallelization:} Excellent potential for parallel Q-value updates
\end{itemize}

 \textbf{4. CPO Constraint-Guided RL}

\textbf{Time Complexity:}
\begin{itemize}
    \item \textbf{Training Phase:} $O(T \times E \times P)$
    \begin{itemize}
        \item $T$ = Number of time steps per episode
        \item $E$ = Number of training episodes
        \item $P$ = Policy optimization complexity (typically $O(n^2)$ for trust region methods)
    \end{itemize}
    \item \textbf{Inference Phase:} $O(T \times P)$
    \item \textbf{Space Complexity:} $O(n^2)$ where $n$ is the policy parameter dimension
\end{itemize}

\textbf{Scalability Characteristics:}
\begin{itemize}
    \item \textbf{Strengths:} Safe policy updates, constraint satisfaction guarantees
    \item \textbf{Limitations:} Conservative updates may slow convergence
    \item \textbf{Memory Usage:} Moderate, scales with policy complexity
    \item \textbf{Parallelization:} Good potential for parallel policy evaluation
\end{itemize}

 \subsection*{C.2 Performance Scaling Analysis}

 \textbf{State Space Scaling}

\begin{table}[htbp]
\centering
\caption{State Space Scaling Characteristics}
\label{tab:state_space_scaling}
\resizebox{0.5\textwidth}{!}{%
\begin{tabular}{lccc}
\toprule
\textbf{Method} & \textbf{Linear Scaling} & \textbf{Memory Growth} & \textbf{Convergence Time} \\
\midrule
Beamsearch & $O(B \times D)$ & $O(B \times D \times S)$ & Exponential \\
Convex & $O(n^3)$ & $O(n^2)$ & Polynomial \\
MDP-ICL & $O(|S| \times |A|)$ & $O(|S| \times |A|)$ & Linear \\
CPO & $O(n^2)$ & $O(n^2)$ & Polynomial \\
\bottomrule
\end{tabular}
}
\end{table}

 \textbf{Constraint Complexity Scaling}

\begin{itemize}
    \item \textbf{Beamsearch:} Linear scaling with number of constraints, but exponential with constraint interaction complexity
    \item \textbf{Convex:} Polynomial scaling with constraint number, efficient for convex constraints
    \item \textbf{MDP-ICL:} Linear scaling with constraint number, but requires constraint encoding in state space
    \item \textbf{CPO:} Linear scaling with constraint number, efficient constraint projection
\end{itemize}

 \subsection*{C.3 Memory and Storage Requirements}

 \textbf{Memory Usage Comparison}

\begin{table}[htbp]
\centering
\caption{Memory Requirements Analysis}
\label{tab:memory_requirements}
\resizebox{0.5\textwidth}{!}{%
\begin{tabular}{lccc}
\toprule
\textbf{Method} & \textbf{Training Memory} & \textbf{Inference Memory} & \textbf{Model Storage} \\
\midrule
Beamsearch & High & High & Low \\
Convex & Medium & Medium & Low \\
MDP-ICL & High & Medium & High \\
CPO & Medium & Low & Medium \\
\bottomrule
\end{tabular}
}
\end{table}

 \textbf{Storage Breakdown:}

\textbf{Beamsearch:}
\begin{itemize}
    \item Training: $O(B \times D \times T \times E)$ trajectory storage
    \item Inference: $O(B \times D \times T)$ active trajectory storage
    \item Model: Policy network parameters only
\end{itemize}

\textbf{Convex:}
\begin{itemize}
    \item Training: $O(n^2)$ optimization matrices
    \item Inference: $O(n^2)$ constraint matrices
    \item Model: Constraint model parameters
\end{itemize}

\textbf{MDP-ICL:}
\begin{itemize}
    \item Training: $O(|S| \times |A|)$ Q-table storage
    \item Inference: $O(|S| \times |A|)$ Q-table access
    \item Model: Complete Q-table or policy network
\end{itemize}

\textbf{CPO:}
\begin{itemize}
    \item Training: $O(n^2)$ policy gradient matrices
    \item Inference: $O(n)$ policy network evaluation
    \item Model: Policy network parameters
\end{itemize}

 \subsection*{C.4 Parallelization Potential}

 \textbf{Parallelization Analysis}

\begin{table}[htbp]
\centering
\caption{Parallelization Characteristics}
\label{tab:parallelization_analysis}
\footnotesize
\begin{tabular}{lccc}
\toprule
\textbf{Method} & \textbf{Train. Parallel} & \textbf{Infer. Parallel} & \textbf{Scalability} \\
\midrule
Beamsearch & Limited & Limited & Low \\
Convex     & High    & High    & High \\
MDP-ICL    & High    & Medium  & High \\
CPO        & High    & High    & High \\
\bottomrule
\end{tabular}

\vspace{1mm}
\begin{minipage}{0.48\textwidth}
\footnotesize
\textbf{Notes:} Train./Infer. Parallel: parallelism during training/inference; Scalability: suitability for large-scale deployment.
\end{minipage}
\end{table}

For large-scale deployments, CO and CPO may provide the best balance of performance, scalability, and constraint satisfaction. 

\appendix

\section*{D. Experiments for Ablation Study}

\subsection*{D.1 Problem and Solution formulations}
For all InD and round dataset we utilise SCP. The first iteration is found without the minimum distance from front car constraint (1e). After this the unit norm of the direction of reference to the front car multiplied by the difference between the ego car and the front car must be larger than some minimum distance. We used 3 iterations of SCP. 

Then there are a few cases to note. In the case that:
\begin{itemize}
    \item the distance between the starting point of lead car and ego car is smaller than the minimum safety distance we discard this case. (There is no feasible solution)
    \item the distance between the ending point of lead car and ego car is smaller than the minimum safety distance we discard this case. (There is no feasible solution)

\end{itemize}
For the highway dataset, we either go left to right or right to left, and so the SCP constraint is replaced by \(s_{x,t} - s^{\mathrm{front}}_{t} \ge d_{\min}\) or \(s_{x,t} - s^{\mathrm{front}}_{t} \le d_{\min}\).

Below are the problem formulations for SCP case.
    
\subsubsection*{Problem Minimise Time}

To find the feasible smallest time we searched through the space  
\([T_{\min},T_{\max}]\) using binary search. If the problem was found infeasible \(T_{\min}\) was increased, if it was feasible \(T_{\max}\) decreased. 
\newtheorem{problem}{Problem}
\begin{problem}[Feasible minimum‑time 2‑D trajectory]\label{prob:ego_plan}
\begin{alignat}{2}
    &\textbf{Given} && \;
      \Delta t,\; T_{\min},T_{\max}\in\mathbb{N},\;
      v_{\max},a_{\max},d_{\min}\in\mathbb{R}_{>0},                                       \notag\\
    &&&(\boldsymbol s_t^{\mathrm{front}})_{t=0}^{T_{\max}-1}\subset\mathbb{R}^2\text{ (lead trajectory)},\notag\\
    &&&\boldsymbol s_0,\boldsymbol v_0,\boldsymbol a_0\in\mathbb{R}^2
      \text{ (initial position, velocity, acceleration)},                             \notag\\
    &&&\boldsymbol s_{\mathrm{goal}}\in\mathbb{R}^2\text{ (goal point)}               \notag
\end{alignat}
Find the smallest horizon \(T^\star\in[T_{\min},T_{\max}]\) and sequences
\(
  (\boldsymbol s_t)_{t=0}^{T^\star},
  (\boldsymbol v_t)_{t=0}^{T^\star},
  (\boldsymbol a_t)_{t=0}^{T^\star}
\subset\mathbb{R}^2
\)
such that, for all \(t=0,\dots,T^\star-1\),
\begin{subequations}\label{eq:scp_constraints}
\begin{align}
  &\boldsymbol s_{t+1}= \boldsymbol s_t + \boldsymbol v_t\,\Delta t, \label{eq:dynamics_s} \\
  &\boldsymbol v_{t+1}= \boldsymbol v_t + \boldsymbol a_t\,\Delta t, \label{eq:dynamics_v} \\
  &\lVert\boldsymbol v_{t}\rVert_2 \le v_{\max},                  \label{eq:speed_limit}\\
  &\lVert\boldsymbol a_{t}\rVert_2 \le a_{\max},                  \label{eq:acc_limit}\\
  &\boldsymbol n_t^\top\!\bigl(\boldsymbol s_t-\boldsymbol s_t^{\mathrm{front}}\bigr)
      \;\ge\; d_{\min},                                            \label{eq:collision}
\end{align}
\end{subequations}
with
\(
  \boldsymbol n_t :=
  \dfrac{\boldsymbol s_t^{\mathrm{ref}}-\boldsymbol s_t^{\mathrm{front}}}
        {\lVert \boldsymbol s_t^{\mathrm{ref}}-\boldsymbol s_t^{\mathrm{front}}\rVert_2}
\)
taken from the reference trajectory of the previous SCP iteration,
and boundary conditions
\[
  \boldsymbol s_0=\boldsymbol s_{\text{init}},\;
  \boldsymbol v_0=\boldsymbol v_{\text{init}},\;
  \boldsymbol a_0=\boldsymbol a_{\text{init}},\qquad
  \boldsymbol s_{T^\star }= \boldsymbol s_{\mathrm{goal}}.
\]
Because only feasibility is sought, the objective function is zero.
\end{problem}

\subsubsection*{Problem Minimise Distance}
\begin{problem}[Minimise Total Travel Distance]\label{prob:travel_distance}
\[
  \min_{\{\boldsymbol s_t\}_{t=0}^{T^\star}} 
      \;\sum_{t=0}^{T^\star-1} 
      \bigl\lVert \boldsymbol s_{t+1}-\boldsymbol s_{t}\bigr\rVert_2
\]
\text{s.t.}\quad same constraints as Problem~\ref{prob:ego_plan}. Here again we use SCP, but there is no need for the binary search as we do not seek to minimise time as well. 
\end{problem}

\subsubsection*{Problem Minimise Effort}
\begin{problem}[Minimise Control Effort (Energy)]\label{prob:control_effort}
\[
  \min_{\{\boldsymbol a_t\}_{t=0}^{T^\star-1}} 
      \;\sum_{t=0}^{T^\star-1} 
      \bigl\lVert \boldsymbol a_{t}\bigr\rVert_2^{2}
\]
\text{s.t.}\quad same constraints as Problem~\ref{prob:ego_plan}.
\end{problem}

\subsubsection*{Problem Minimise Jerk}
\begin{problem}[Minimise Jerk (Comfort)]\label{prob:jerk}
\[
  \min_{\{\boldsymbol a_t\}_{t=1}^{T^\star-1}} 
      \;\sum_{t=1}^{T^\star-1} 
      \bigl\lVert \boldsymbol a_{t+1}-\boldsymbol a_{t}\bigr\rVert_2^{2}
\]
\text{s.t.}\quad same constraints as Problem~\ref{prob:ego_plan}.
\end{problem}

\subsection*{D.2 Algorithms for Convex Planner}

\begin{algorithm}[H]
\caption{Minimize Objective with SCP (for inD/rounD datasets)}
\label{alg:min_obj_scp}
\begin{algorithmic}[1]
\Function{MinimiseObjectiveSCP}{$s_{\text{init}}, v_{\text{init}}, a_{\text{init}}, s_{\text{front}}, s_{\text{goal}}, T, \dots$}
    \State $s_{\text{ref}} \gets \text{None}$
    \For{$i \gets 1$ to $N_{\text{scp}}$}
        \State Define convex optimization problem $\mathcal{P}$:
        \State \textbf{Objective}: Minimize Distance, Effort, or Jerk.
        \State \textbf{Constraints}:
        \State \quad $s_0, v_0, a_0 = s_{\text{init}}, v_{\text{init}}, a_{\text{init}}$
        \State \quad $s_{T-1} = s_{\text{goal}}$
        \For{$t \gets 0$ to $T-2$}
            \State $s_{t+1} = s_t + v_t \Delta t, \quad v_{t+1} = v_t + a_t \Delta t$
            \State $\|v_{t+1}\|_2 \le v_{\max}, \quad \|a_{t+1}\|_2 \le a_{\max}$
            \If{$s_{\text{ref}}$ is not None}
                \State $\Delta s \gets s_{\text{ref}, t} - s_{\text{front}, t}$
                \If{$\|\Delta s\|_2 > \epsilon$}
                    \State $n_t \gets \Delta s / \|\Delta s\|_2$
                    \State $n_t^T (s_t - s_{\text{front}, t}) \ge d_{\min}$ \Comment{Linearized collision avoidance}
                \EndIf
            \EndIf
        \EndFor
        \State Solve $\mathcal{P}$ to get trajectory $(s^*, v^*, a^*)$.
        \If{Problem is infeasible}
            \State \Return None \Comment{SCP failed}
        \EndIf
        \State $s_{\text{ref}} \gets s^*$ \Comment{Update reference for next iteration}
    \EndFor
    \State \Return $(T, s^*, v^*, a^*)$
\EndFunction
\end{algorithmic}
\end{algorithm}

\begin{algorithm}[H]
\caption{Minimize Time with SCP (for inD/rounD datasets)}
\label{alg:min_time_scp}
\begin{algorithmic}[1]
\Function{MinimiseTimeSCP}{$s_{\text{init}}, \dots, T_{\min}, T_{\max}, N_{\text{scp}}, \dots$}
    \State $left \gets T_{\min}, \quad right \gets T_{\max}$
    \State $best\_solution \gets \text{None}$
    \While{$left \le right$}
        \State $T \gets (left + right) // 2$
        \State $s_{\text{ref}} \gets \text{None}$
        \State $is\_feasible\_for\_T \gets \text{False}$
        \For{$i \gets 1$ to $N_{\text{scp}}$} \Comment{Inner SCP loop for feasibility}
            \State Define feasibility problem $\mathcal{P}_{\text{feas}}$ with horizon $T$ (as in Alg \ref{alg:min_obj_scp}, but with objective `minimize 0`).
            \State Solve $\mathcal{P}_{\text{feas}}$ to get $(s^*, v^*, a^*)$.
            \If{Problem is feasible}
                \State $s_{\text{ref}} \gets s^*$
                \If{$i = N_{\text{scp}}$}
                    \State $is\_feasible\_for\_T \gets \text{True}$
                \EndIf
            \Else
                \State $is\_feasible\_for\_T \gets \text{False}$
                \State \textbf{break} \Comment{SCP iteration failed}
            \EndIf
        \EndFor
        \If{$is\_feasible\_for\_T$}
            \State $best\_solution \gets (T, s^*, v^*, a^*)$
            \State $right \gets T - 1$ \Comment{Feasible, try a shorter time}
        \Else
            \State $left \gets T + 1$ \Comment{Infeasible, need more time}
        \EndIf
    \EndWhile
    \State \Return $best\_solution$
\EndFunction
\end{algorithmic}
\end{algorithm}

\begin{algorithm}[H]
\caption{Minimize Objective (for highD dataset)}
\label{alg:min_obj_highd}
\begin{algorithmic}[1]
\Function{MinimiseObjectiveHighD}{$s_{\text{init}}, v_{\text{init}}, a_{\text{init}}, s_{\text{front}}, s_{\text{goal}}, T, \dots$}
    \State Define convex optimization problem $\mathcal{P}$:
    \State \textbf{Objective}: Minimize Distance, Effort, or Jerk.
    \State \textbf{Constraints}:
    \State \quad $s_0, v_0, a_0 = s_{\text{init}}, v_{\text{init}}, a_{\text{init}}$
    \State \quad $s_{T-1} = s_{\text{goal}}$
    \For{$t \gets 0$ to $T-2$}
        \State $s_{t+1} = s_t + v_t \Delta t, \quad v_{t+1} = v_t + a_t \Delta t$
        \State $\|v_{t+1}\|_2 \le v_{\max}, \quad \|a_{t+1}\|_2 \le a_{\max}$
    \EndFor
    \If{$s_{\text{goal},x} \ge s_{\text{init},x}$} \Comment{Ego vehicle is behind or moving right}
        \For{$t \gets 0$ to $T-1$}
             \State $s_{x, t} \le s_{\text{front}, x, t} - d_{\min}$ \Comment{Simplified collision avoidance}
        \EndFor
    \Else \Comment{Ego vehicle is ahead or moving left}
        \For{$t \gets 0$ to $T-1$}
             \State $s_{x, t} \ge s_{\text{front}, x, t} + d_{\min}$
        \EndFor
    \EndIf
    \State Solve $\mathcal{P}$ to get trajectory $(s^*, v^*, a^*)$.
    \If{Problem is feasible}
        \State \Return $(T, s^*, v^*, a^*)$
    \Else
        \State \Return None
    \EndIf
\EndFunction
\end{algorithmic}
\end{algorithm}

\begin{algorithm}[H]
\caption{Minimize Time (for highD dataset)}
\label{alg:min_time_highd}
\begin{algorithmic}[1]
\Function{MinimiseTimeHighD}{$s_{\text{init}}, \dots, T_{\min}, T_{\max}, \dots$}
    \State $left \gets T_{\min}, \quad right \gets T_{\max}$
    \State $best\_solution \gets \text{None}$
    \While{$left \le right$}
        \State $T \gets (left + right) // 2$
        \State Define feasibility problem $\mathcal{P}_{\text{feas}}$ with horizon $T$ (as in Alg \ref{alg:min_obj_highd}, but with objective `minimize 0`).
        \State Solve $\mathcal{P}_{\text{feas}}$ to get $(s^*, v^*, a^*)$.
        \If{Problem is feasible}
            \State $best\_solution \gets (T, s^*, v^*, a^*)$
            \State $right \gets T - 1$ \Comment{Feasible, try a shorter time}
        \Else
            \State $left \gets T + 1$ \Comment{Infeasible, need more time}
        \EndIf
    \EndWhile
    \State \Return $best\_solution$
\EndFunction
\end{algorithmic}
\end{algorithm}
Here $\mathcal{P}$ defines the convex problems as earlier described.

\subsection*{D.3 Results}
Here are the results. Tracks is the number of ego and lead car pairs tested. Inf. relates to the number of pairs that the solver found infeasible. Bad denotes the number of cases that were deemed infeasible through checks such as the starting of the front car and the ego car being within minimum safety distance. Avg, std, Min, Max relates to the compute time taken to determine a solution in seconds (this include these checks).

The vast majority of the infeasible problems are found as infeasible due to i) The Lead Car and the Ego car heading in opposing directions such that for somewhere in the middle of the run the minimum safety distance is breached. The three round SCP with the first naive estimate of no collision constraints is not always able to solve this. ii) The starting point of the front car  only just passes the d\_min threshold and when the ego car is initialized with some starting velocity or acceleration the d\_min and physics constraints make the problem infeasible.


\begin{table}[H]
\centering
\caption{Objective: Time (Part 1)}
\label{tab:time_part1}
\begin{tabular}{
  l
  S[table-format=4]
  S[table-format=3] @{\, (} S[table-format=2.1] @{) \,}
  S[table-format=3] @{\, (} S[table-format=2.1] @{) \,}
}
\toprule
Dataset & {\#Tracks} &
\multicolumn{2}{c}{Inf.\,(\%)} &
\multicolumn{2}{c}{Bad.\,(\%)} \\ \midrule
inD01    &  377 &  24 &  6.4 &  0 &  0.0 \\
roundD01 &  260 &  50 & 19.2 &  0 &  0.0 \\
highD01  & 1018 & 161 & 15.8 & 42 &  4.1 \\
\bottomrule
\end{tabular}
\end{table}

\begin{table}[H]
\centering
\caption[]{Objective: Time (Part 2, continued from Table~\ref{tab:time_part1})}
\begin{tabular}{
  l
  S[table-format=2.2]
  S[table-format=2.2]
  S[table-format=2.2]
  S[table-format=3.2]
}
\toprule
Dataset & {Avg} & {Std} & {Min} & {Max} \\
\midrule
inD01    & 12.89 &  9.42 & 1.07 &  96.98 \\
roundD01 & 22.94 & 13.45 & 4.26 & 107.35 \\
highD01  & 10.16 &  3.54 & 0.44 &  26.65 \\
\bottomrule
\end{tabular}
\end{table}

\begin{table}[H]
\centering
\caption{Objective: Distance (Part 1)}
\label{tab:distance_part1}
\begin{tabular}{
  l
  S[table-format=4]
  S[table-format=3] @{\, (} S[table-format=2.1] @{) \,}
  S[table-format=3] @{\, (} S[table-format=2.1] @{) \,}
}
\toprule
Dataset & {\#Tracks} &
\multicolumn{2}{c}{Inf.\,(\%)} &
\multicolumn{2}{c}{Bad.\,(\%)} \\ \midrule
inD01    &  377 &  33 &  8.8 &  0 &  0.0 \\
roundD01 &  260 &  64 & 24.6 &  0 &  0.0 \\
highD01  & 1018 & 160 & 15.7 & 42 &  4.1 \\
\bottomrule
\end{tabular}
\end{table}

\begin{table}[H]
\centering
\caption[]{Objective: Distance (Part 2, continued from Table~\ref{tab:distance_part1})}
\begin{tabular}{
  l
  S[table-format=2.2]
  S[table-format=2.2]
  S[table-format=2.2]
  S[table-format=2.2]
}
\toprule
Dataset & {Avg} & {Std} & {Min} & {Max} \\
\midrule
inD01    &  6.20 &  8.96 & 0.43 & 60.08 \\
roundD01 & 12.58 & 10.16 & 1.93 & 96.29 \\
highD01  &  2.55 &  0.98 & 0.14 &  8.45 \\
\bottomrule
\end{tabular}
\end{table}
\begin{table}[H]
\centering
\caption{Objective: Effort (Part 1)}
\label{tab:effort_part1}
\begin{tabular}{
  l
  S[table-format=4]
  S[table-format=3] @{\, (} S[table-format=2.1] @{) \,}
  S[table-format=3] @{\, (} S[table-format=2.1] @{) \,}
}
\toprule
Dataset & {\#Tracks} &
\multicolumn{2}{c}{Inf.\,(\%)} &
\multicolumn{2}{c}{Bad.\,(\%)} \\ \midrule
inD01    &  377 &  30 &  8.0 &  0 &  0.0 \\
roundD01 &  260 &  60 & 23.1 &  0 &  0.0 \\
highD01  & 1018 & 160 & 15.7 & 42 &  4.1 \\
\bottomrule
\end{tabular}
\end{table}

\begin{table}[H]
\centering
\caption[]{Objective: Effort (Part 2, continued from Table~\ref{tab:effort_part1})}
\begin{tabular}{
  l
  S[table-format=2.2]
  S[table-format=2.2]
  S[table-format=2.2]
  S[table-format=2.2]
}
\toprule
Dataset & {Avg} & {Std} & {Min} & {Max} \\
\midrule
inD01    &  7.77 & 11.60 & 0.71 & 81.38 \\
roundD01 &  9.96 &  7.88 & 1.62 & 79.38 \\
highD01  &  1.44 &  0.42 & 0.09 &  4.16 \\
\bottomrule
\end{tabular}
\end{table}

\begin{table}[H]
\centering
\caption{Objective: Jerk (Part 1)}
\label{tab:jerk_part1}
\begin{tabular}{
  l
  S[table-format=4]
  S[table-format=3] @{\, (} S[table-format=2.1] @{) \,}
  S[table-format=3] @{\, (} S[table-format=2.1] @{) \,}
}
\toprule
Dataset & {\#Tracks} &
\multicolumn{2}{c}{Inf.\,(\%)} &
\multicolumn{2}{c}{Bad.\,(\%)} \\ \midrule
inD01    &  377 &  28 &  7.4 &  0 &  0.0 \\
roundD01 &  260 &  63 & 24.2 &  0 &  0.0 \\
highD01  & 1018 & 160 & 15.7 & 42 &  4.1 \\
\bottomrule
\end{tabular}
\end{table}

\begin{table}[H]
\centering
\caption[]{Objective: Jerk (Part 2, continued from Table~\ref{tab:jerk_part1})}
\begin{tabular}{
  l
  S[table-format=2.2]
  S[table-format=2.2]
  S[table-format=2.2]
  S[table-format=2.2]
}
\toprule
Dataset & {Avg} & {Std} & {Min} & {Max} \\
\midrule
inD01    &  5.58 &  7.96 & 0.42 & 62.32 \\
roundD01 & 10.03 &  8.25 & 1.83 & 76.70 \\
highD01  &  1.54 &  0.50 & 0.14 &  4.52 \\
\bottomrule
\end{tabular}
\end{table}


For the Round dataset there is are two additonal infeasbile situation: iii) The front car passses through the goal state ending at some point only just meeting the d\_min criteria, and so only giving access to the goal region for a short period of time. iV) The roundabout dataset features many low speeds and tight, continuous turns. This higher variability in speed and high-curvature geometry cause the the first linear SCP choice to be even worse, making it difficult for the iterative SCP solver to handle, causing it to fail more often than on straighter roads. 

Finally, for the highD dataset, another type of infeasible case arises. In these runs, overtaking was not permitted due to the $d_{\min}$ constraints being formulated as hard hyperplanes perpendicular to the $x$-axis, effectively enforcing longitudinal safety buffers. Consequently, when the lead vehicle's stopping point is ahead of the ego vehicle's target, the ego is unable to proceed without violating the minimum distance constraint. In such scenarios, especially under lane-keeping conditions, the planner fails to generate a feasible trajectory, resulting in early termination of the optimization.

\subsection*{D.4 Example Runs}

\subsubsection*{Infeasible (inD dataset)} \mbox{}\\

This example illustrates a failure case where the ego vehicle (Real Car ID 26) violates the minimum safety distance constraint from the lead vehicle (ID 22). As shown in the bottom-right subplot, the inter-vehicle distance drops below the $d_{\min} = 10$\,m threshold for an extended duration. Despite satisfying dynamic constraints, the optimization problem becomes infeasible due to sustained collision risk, especially in scenarios with similar trajectories and limited overtaking space.

\begin{figure}[H]
    \centerline{\includegraphics[width=\columnwidth]{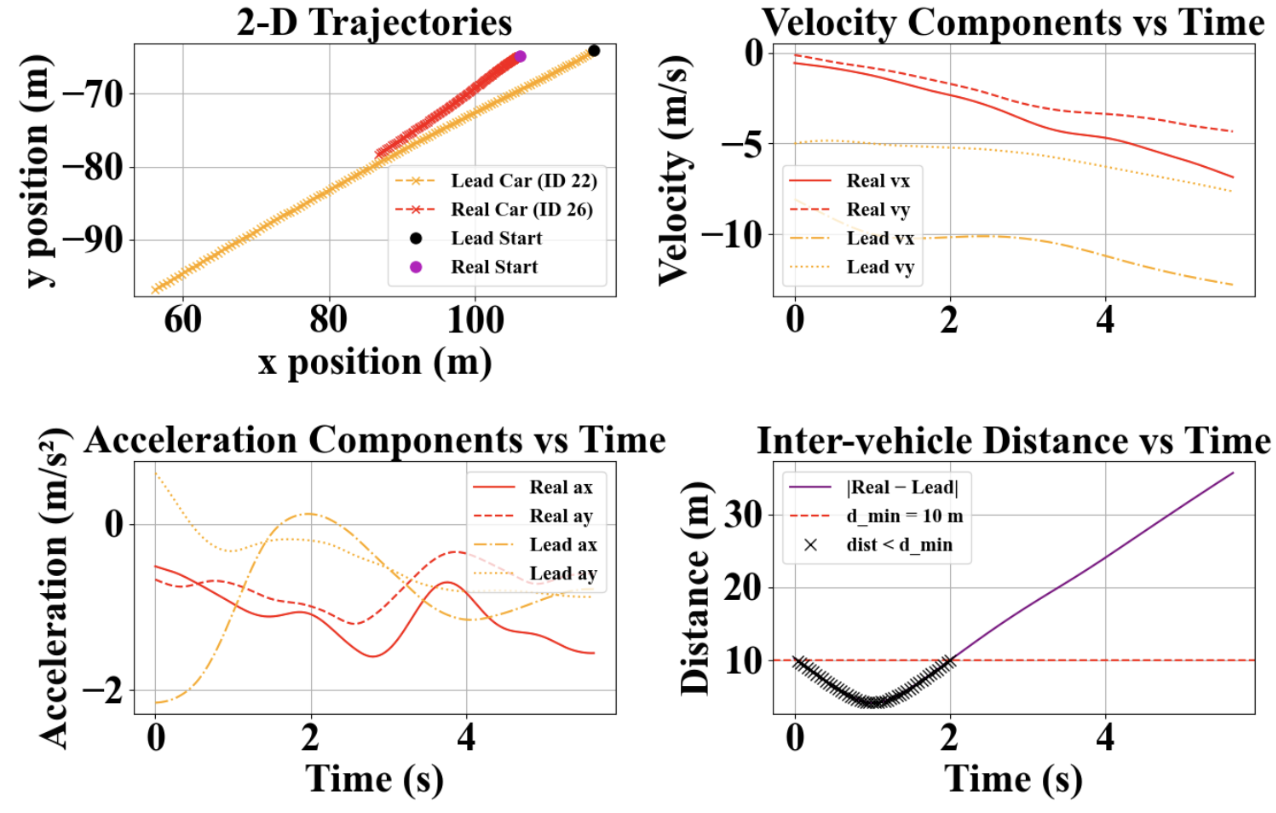}}
    \caption{Real id: 26, Front Id: 22, Infeasible}
    \label{fig:placeholder}
\end{figure}

\subsubsection*{Feasible Turn Right (inD dataset)} 

\mbox{}\\

\begin{figure}[htbp]
    \centering
    \includegraphics[width=1\linewidth]{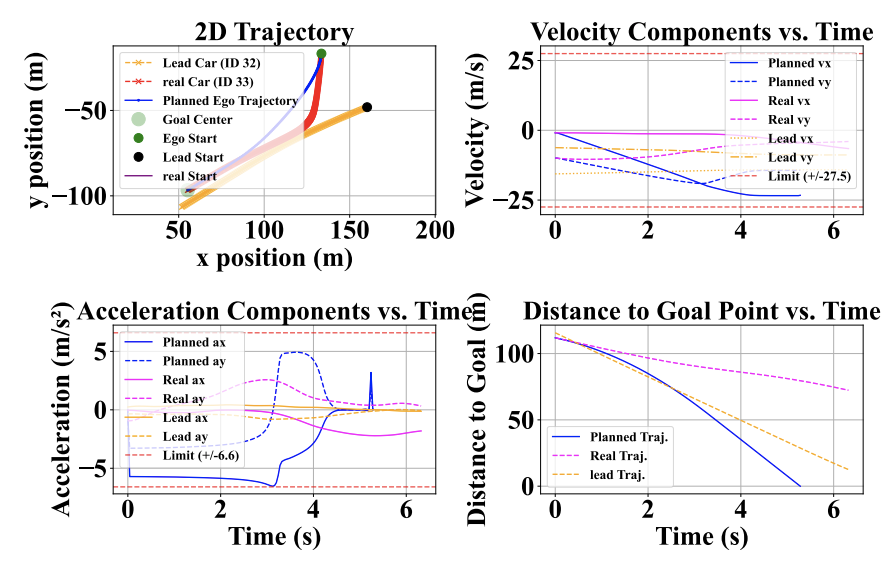}
    \caption{Ego 33, Front 32 Objective: Time Compute Time Taken: 8.91s Planned Steps: 133}
    \label{fig:placeholder}
\end{figure}

\begin{figure}[htbp]
    \centerline{\includegraphics[width=\columnwidth]{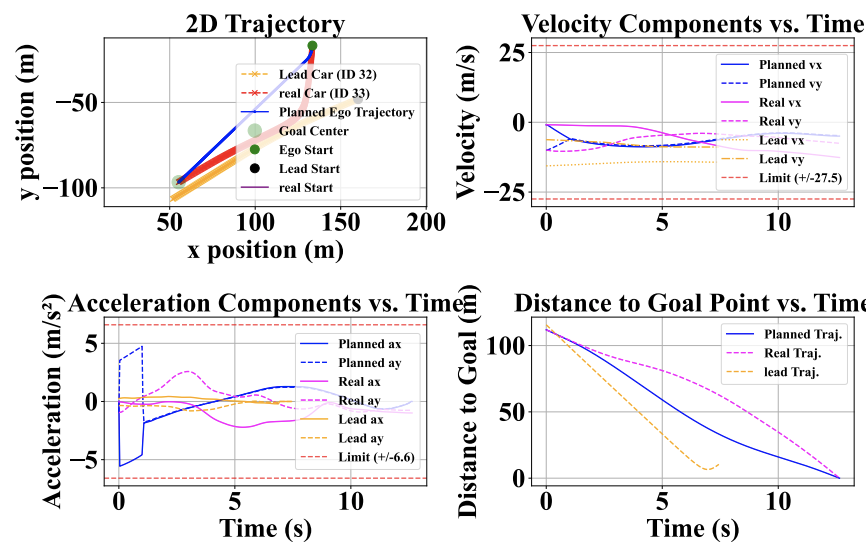}}
    \caption{Ego 33, Front 32 Objective: Distance Compute Time Taken: 5.35s Planned Steps: 317}
    
    \label{fig:ind_R_D}
\end{figure}

\begin{figure}[htbp]
    \centerline{\includegraphics[width=\columnwidth]{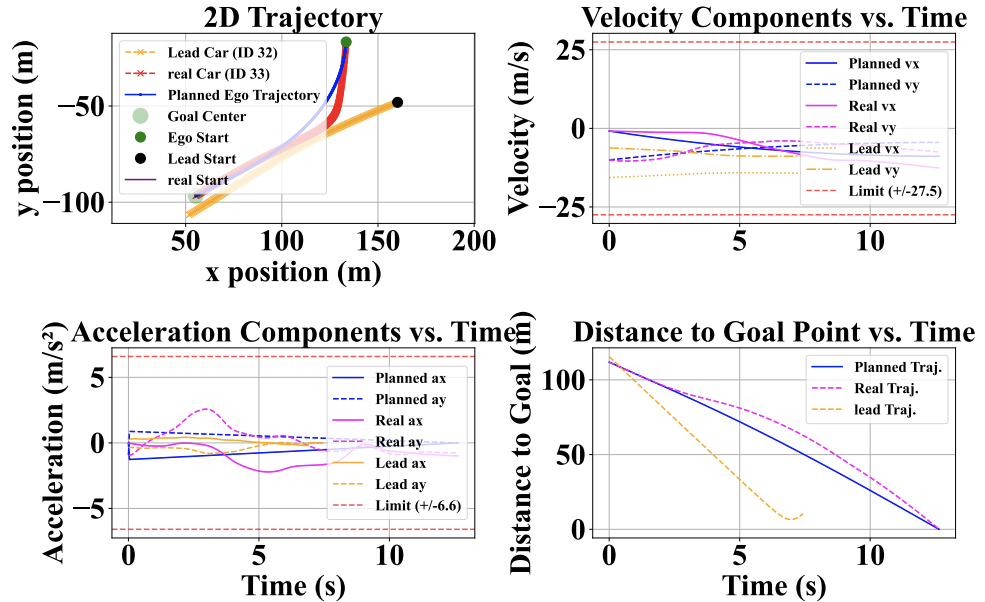}}
    \caption{Ego 33, Front 32 Objective: Effort Compute Time Taken: 4.22s Planned Steps: 317}
    
    \label{fig:ind_r_E}
\end{figure}

\begin{figure}[htbp]
    \centerline{\includegraphics[width=\columnwidth]{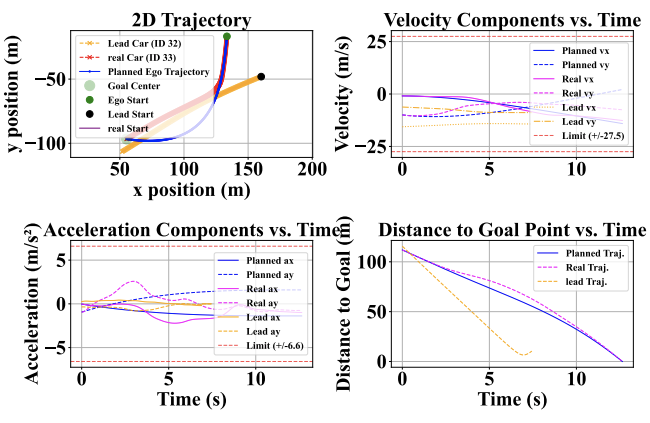}}
    \caption{Ego 33, Front 32 Objective: Jerk Compute Time Taken: 4.27s Planned Steps: 317}
    
    \label{fig:ind_r_J}
\end{figure}

\newpage
\subsubsection*{Feasible Turn Left (inD dataset)} \mbox{}\\

\begin{figure}[H]
    \centerline{\includegraphics[width=\columnwidth]{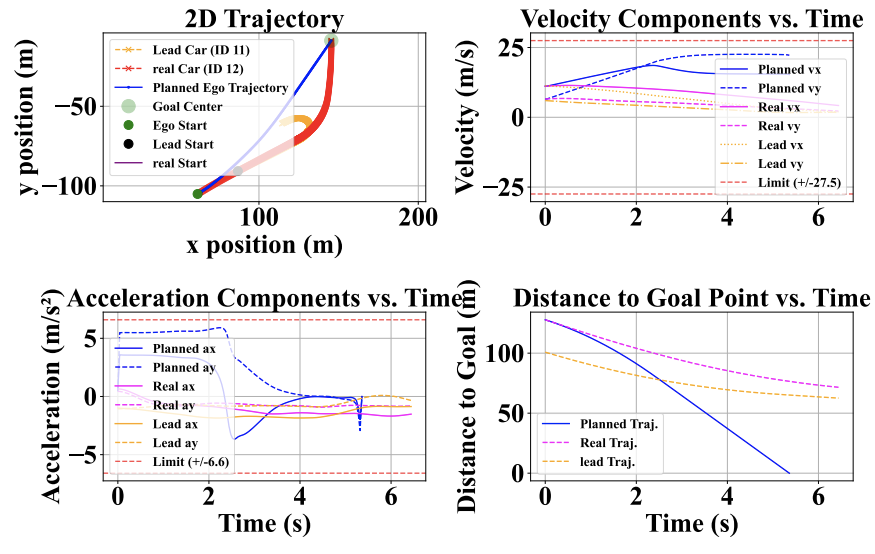}}
    \caption{Ego 12, Front 11 Objective: Time Compute Time Taken: 10.07s Planned Steps: 135}
    
    \label{fig:ind_l_t}
\end{figure}

\begin{figure}[H]
    \centerline{\includegraphics[width=\columnwidth]{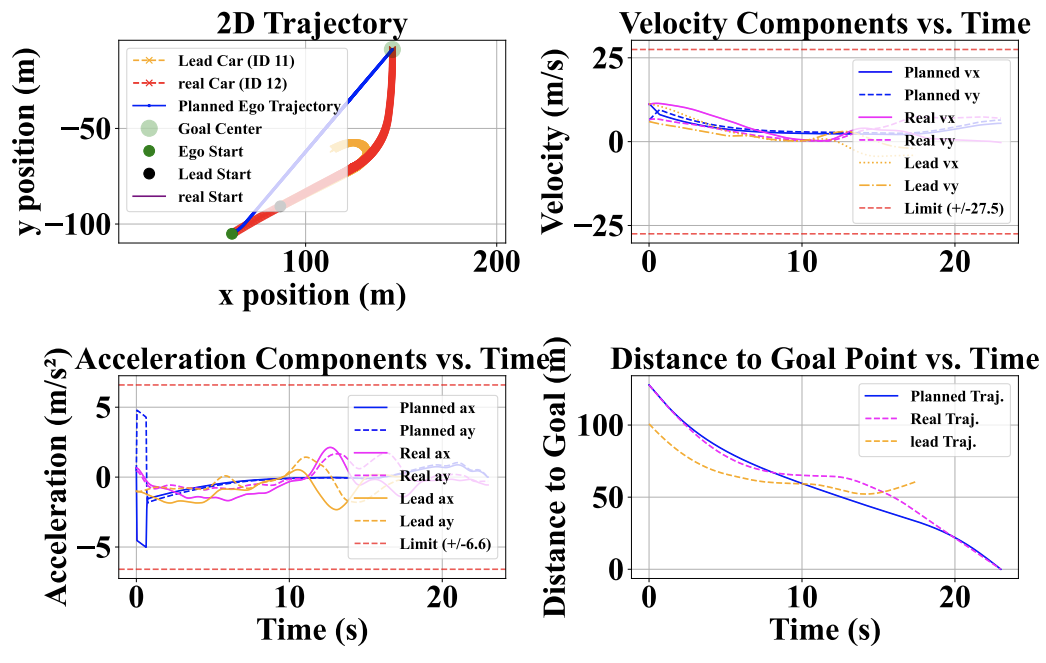}}
    
    \caption{Ego 12, Front 11 Objective: Distance Compute Time Taken: 11.32s Planned Steps: 576}
    \label{fig:i_l_d}
\end{figure}

\begin{figure}[H]
    \centerline{\includegraphics[width=\columnwidth]{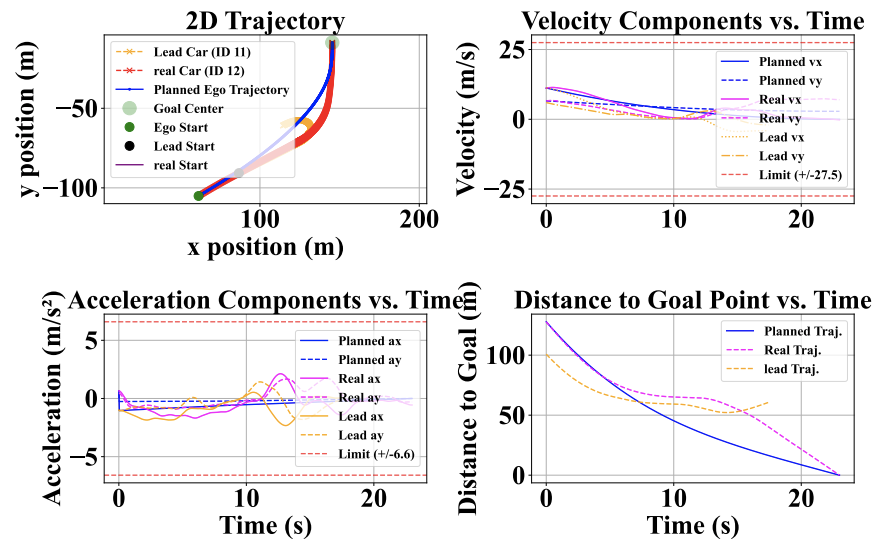}}
    \caption{Ego 12, Front 11 Objective: Effort Compute Time Taken: 8.57s Planned Steps: 576}
    
    \label{fig:ind_L_e}
\end{figure}

\begin{figure}[H]
    \centerline{\includegraphics[width=\columnwidth]{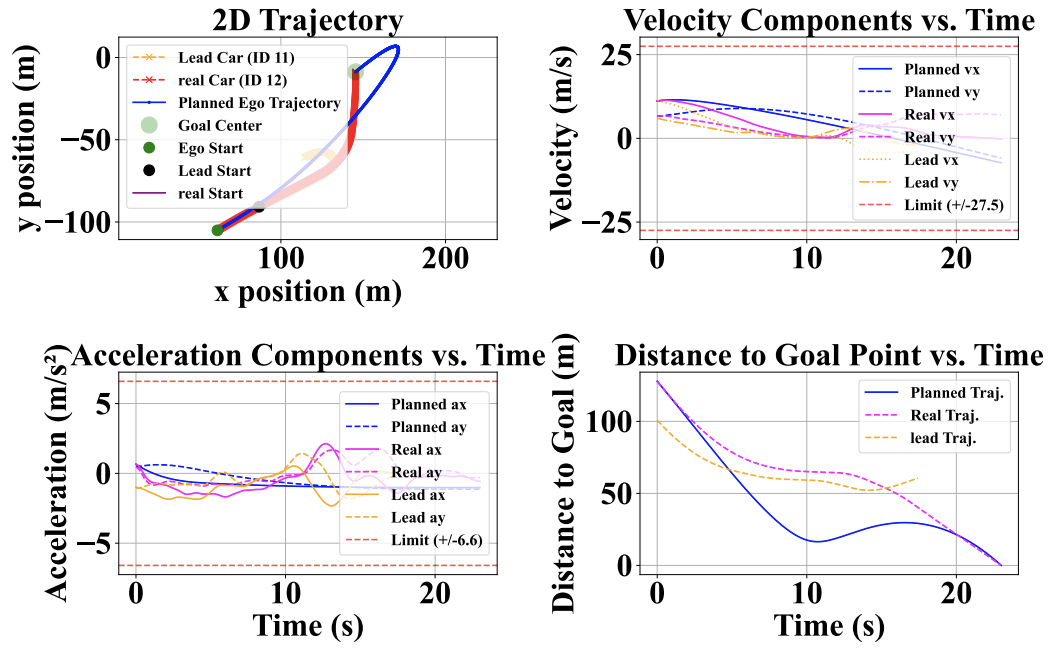}}
    \caption{Ego 12, Front 11 Objective: Jerk Compute Time Taken: 8.44s Planned Steps: 576}
    
    \label{fig:INd_L_j}
\end{figure}

\subsubsection*{Feasible Big Turn (Round Dataset)} \mbox{}\\

\begin{figure}[H]
    \centerline{\includegraphics[width=\columnwidth]{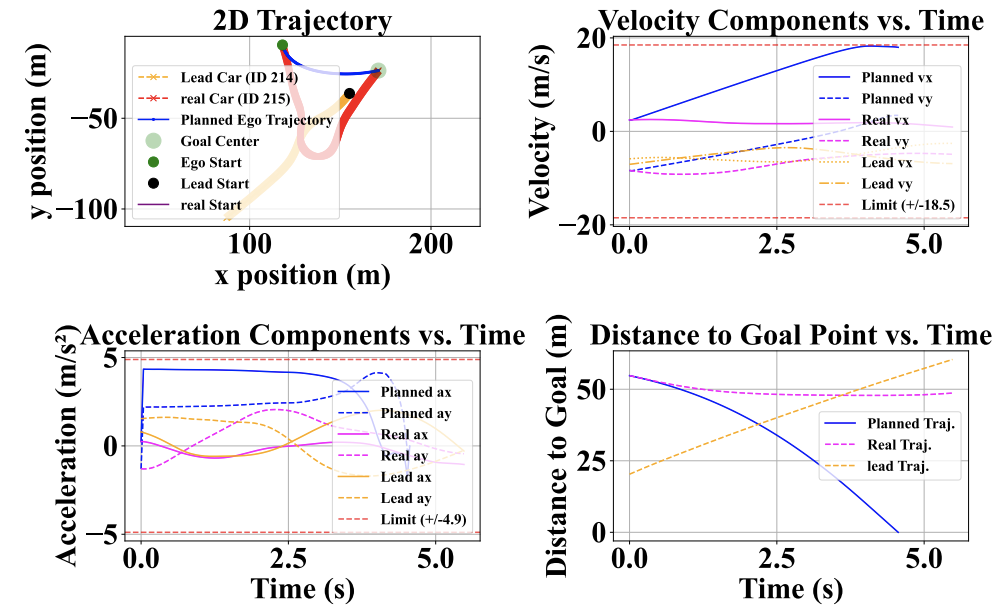}}
    \caption{Ego 215, Front 214 Objective: Time Compute Time Taken: 10.66s Planned Steps: 115}
    \label{fig:round_big_T}
\end{figure}

\begin{figure}[H]
    \centerline{\includegraphics[width=\columnwidth]{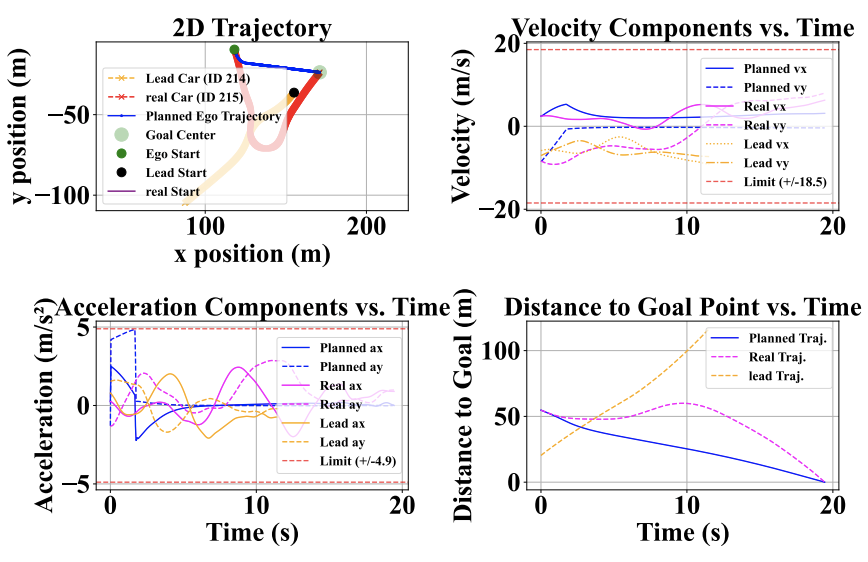}}
    \caption{Ego 215, Front 214, Objective: Distance, Compute Time Taken: 9.30s, Planned Steps: 487}
    
    \label{fig:round_big_D}
\end{figure}

\begin{figure}[H]
    \centerline{\includegraphics[width=\columnwidth]{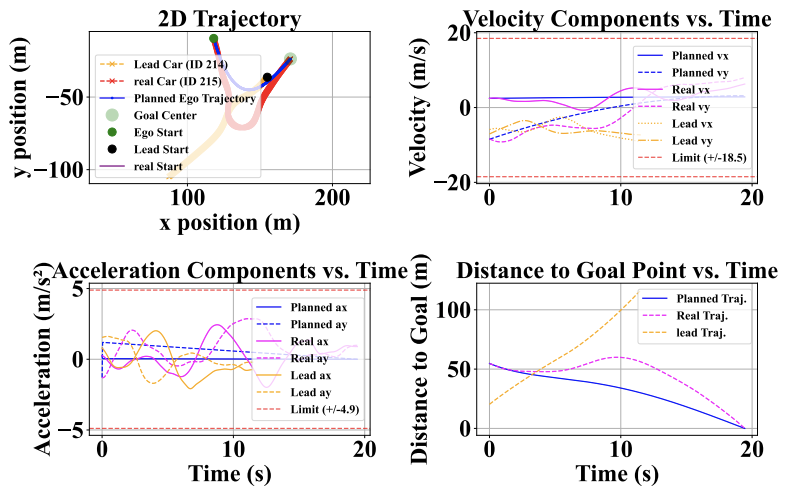}}
    \caption{Ego 215, Front 214, Objective: Effort, Compute Time Taken: 8.51s, Planned Steps: 487}
    
    \label{fig:round_big_E}
\end{figure}

\begin{figure}[H]
    \centerline{\includegraphics[width=\columnwidth]{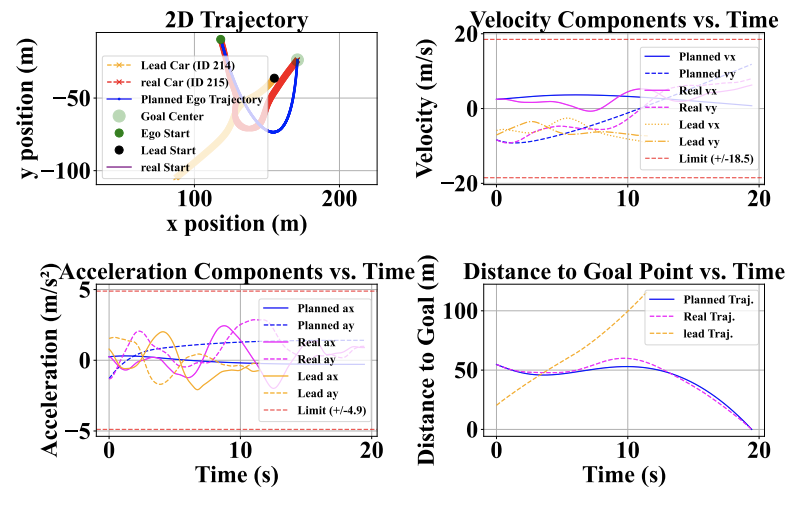}}
    \caption{Ego 215, Front 214, Objective: Jerk, Compute Time Taken: 8.83s, Planned Steps: 487}
    
    \label{fig:round_big_J}
\end{figure}

\subsubsection*{Infeasible (High Dataset)} \mbox{}\\

\begin{figure}[H]
    \centerline{\includegraphics[width=\columnwidth]{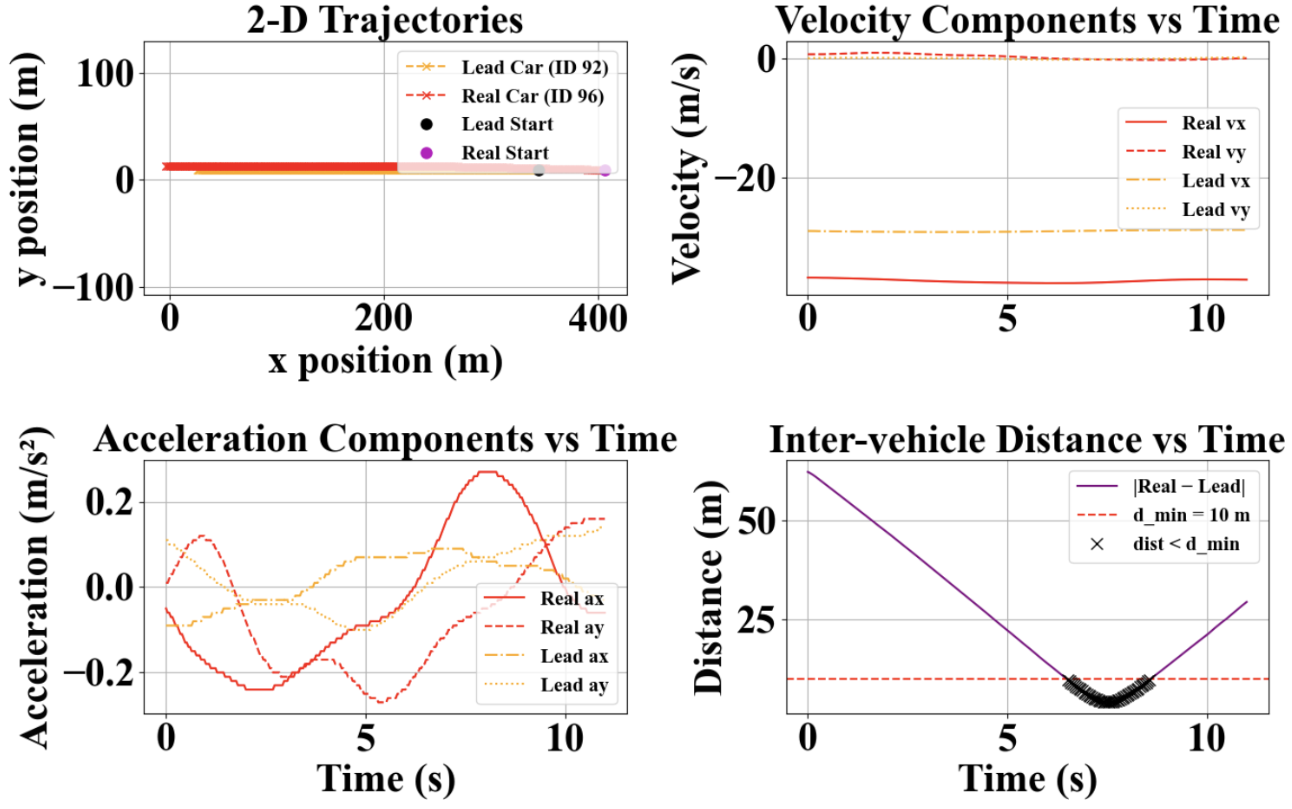}}
    \caption{Infeasible, Real:96, Lead:92, Objective = Time}
    \label{fig:high_infeas}
\end{figure}


\newpage
\subsubsection*{Feasible (High Dataset)} \mbox{}\\

\begin{figure}[H]
    \centerline{\includegraphics[width=\columnwidth]{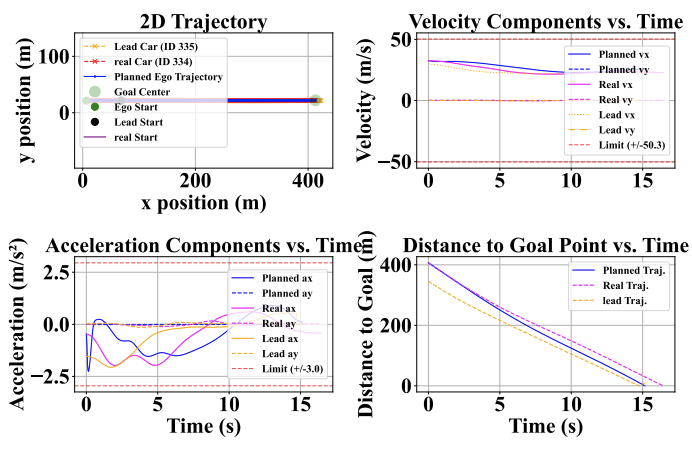}}
    \caption{Ego 334, Front 335, Objective: Time, Compute Time Taken: 13.33s Planned Steps: 380}
    \label{fig:high_f_T}
\end{figure}

\begin{figure}[H]
    \centerline{\includegraphics[width=\columnwidth]{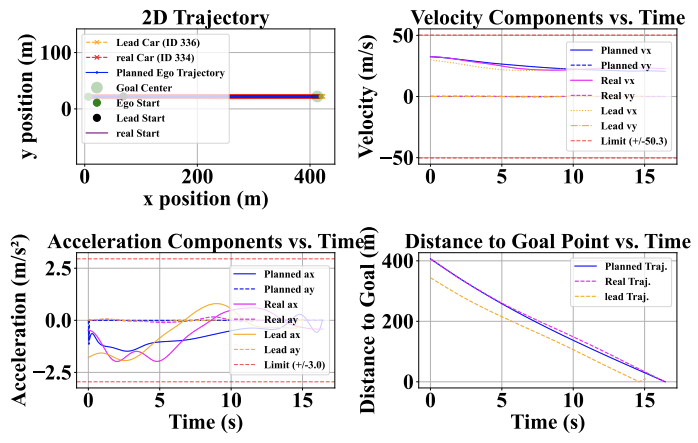}}
    \caption{Ego 334, Front 336, Objective: Distance, Compute Time Taken: 2.15s Planned Steps: 412}
    
    \label{fig:high_f_D}
\end{figure}

\begin{figure}[H]
    \centerline{\includegraphics[width=\columnwidth]{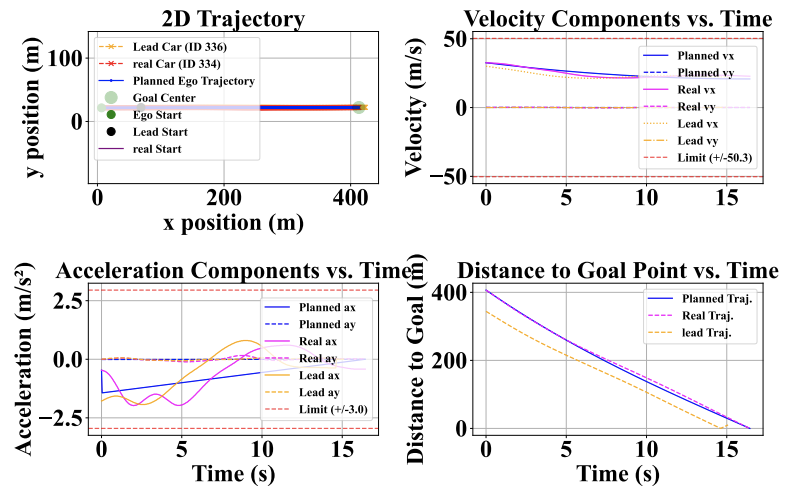}}
    \caption{Ego 334, Front 336, Objective: Effort, Compute Time Taken: 1.76s, Planned Steps: 412}
    
    \label{fig:high_F_E}
\end{figure}

\begin{figure}[H]
    \centerline{\includegraphics[width=\columnwidth]{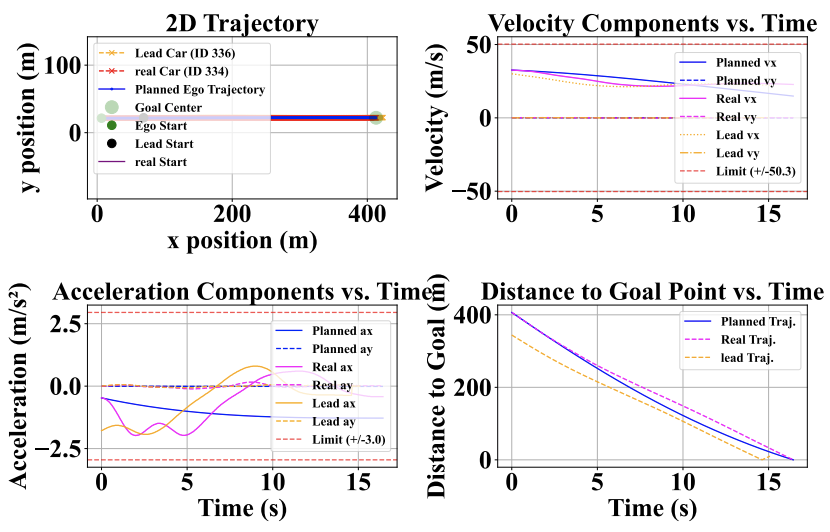}}
    \caption{Ego 334, Front 336, Objective: Jerk, Compute Time Taken: 1.72s, Planned Steps: 412}
    
    \label{fig:High_F_J}
\end{figure}

\section*{Trajectory Sampling, Generation, and Visualization (MDP, BS, CPO)}

\begin{figure}[htbp]
    \centering
    \includegraphics[width=0.5\textwidth]{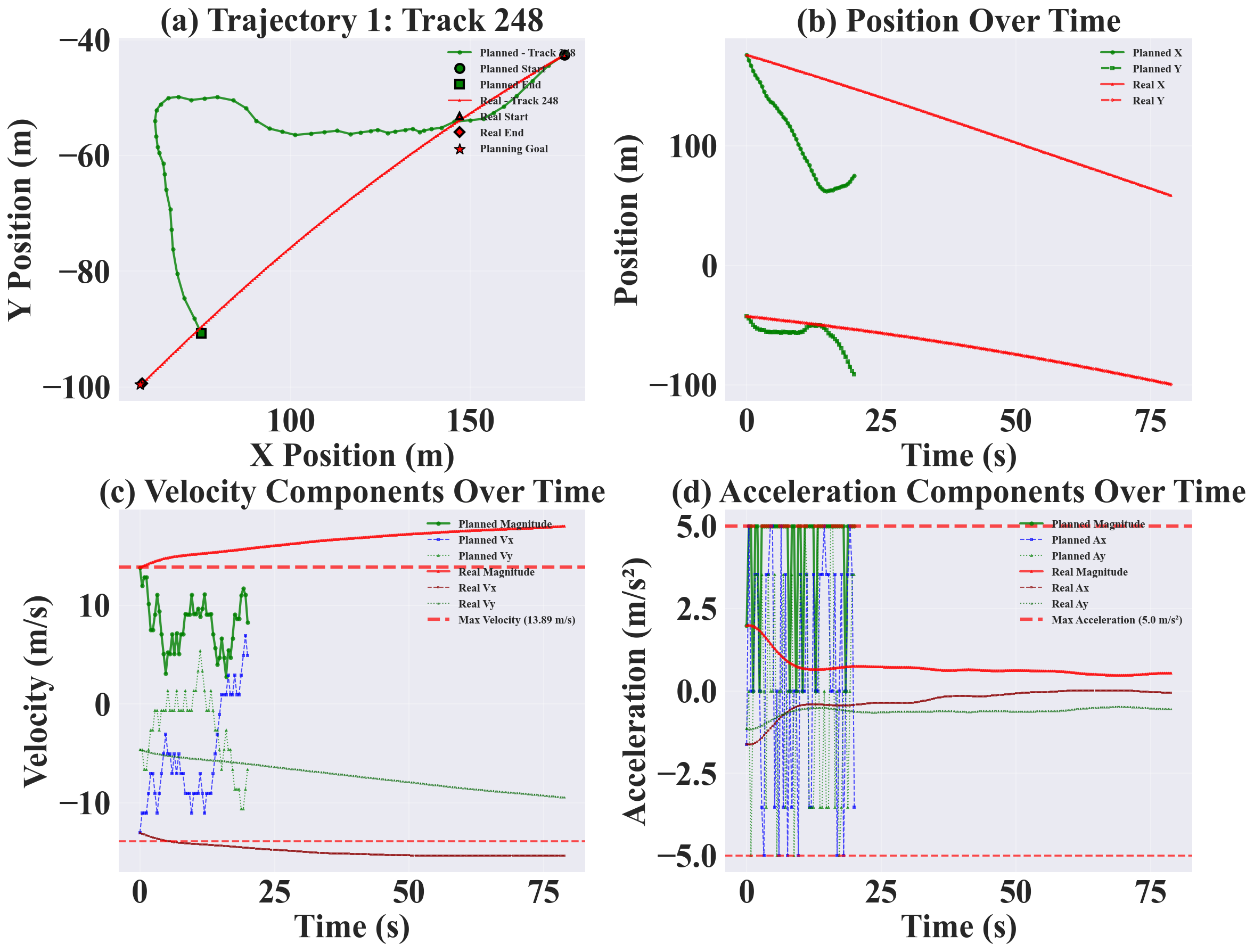}
    \caption{Illustration of constraint and trajectory (MDP).Reward Function: Progress to goal + Hard constraint penalties (velocity$\leq$13.89m/s, acceleration$\leq$5m/s², collision avoidance) + Soft constraint bonuses + Time penalty. Reward: -2416.13, Final Velocity: 13.68 m/s, Final Acceleration: 5.00 m/s², Distance Traveled: 121.77 m}
    \label{fig:constraint_diagram}
\end{figure}

\begin{figure}[htbp]
    \centering
    \includegraphics[width=0.5\textwidth]{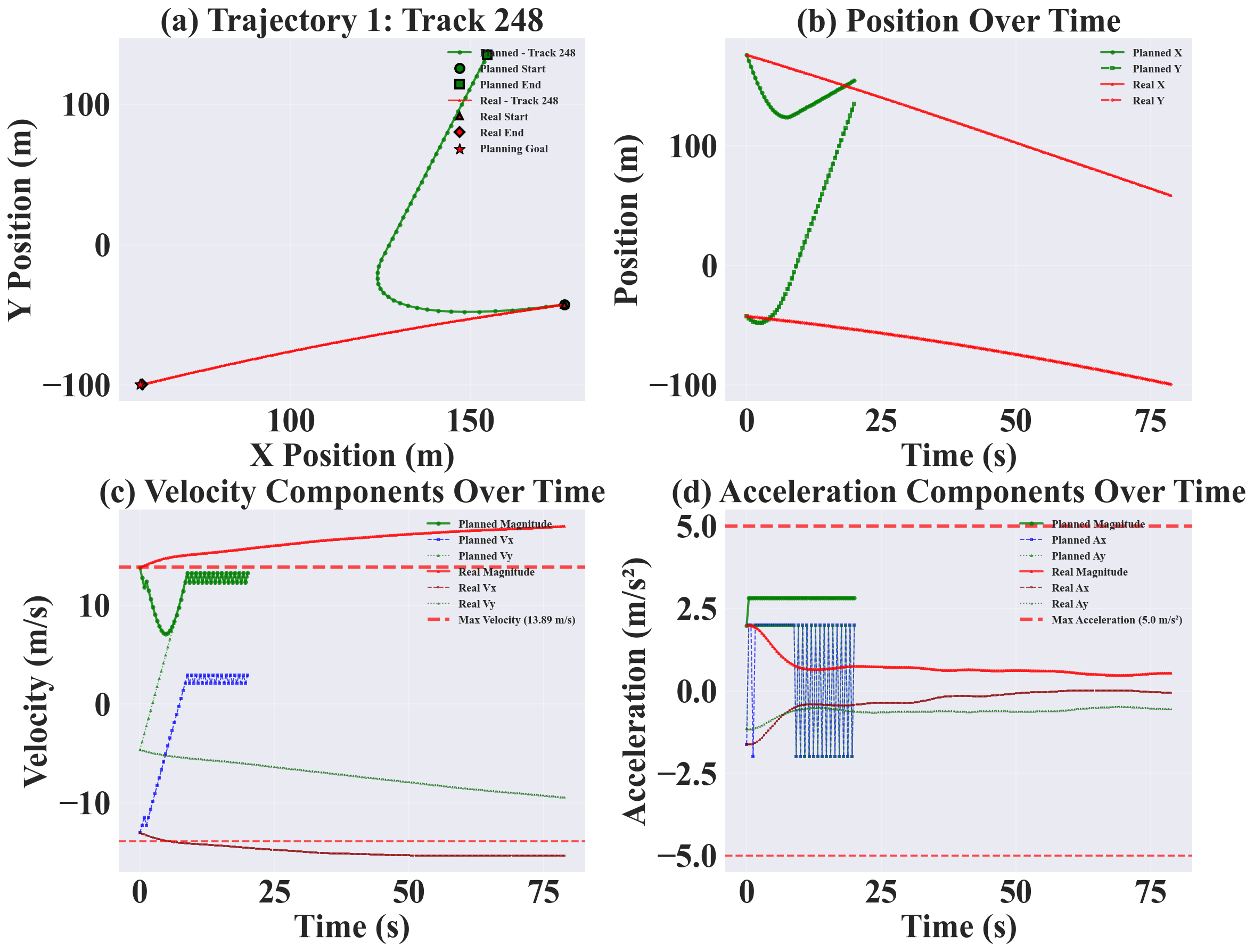}
    \caption{Illustration of constraint and trajectory (BS). Reward Function: Progress to goal + Hard constraint penalties (velocity$\leq$13.89m/s, acceleration$\leq$5m/s², collision avoidance) + Soft constraint bonuses + Time penalty. Reward: -9424.76, Final Velocity: 12.67 m/s, Final Acceleration: 2.83 m/s², Distance Traveled: 178.18 m}
    \label{fig:constraint_diagram}
\end{figure}

\begin{figure}[htbp]
    \centering
    \includegraphics[width=0.5\textwidth]{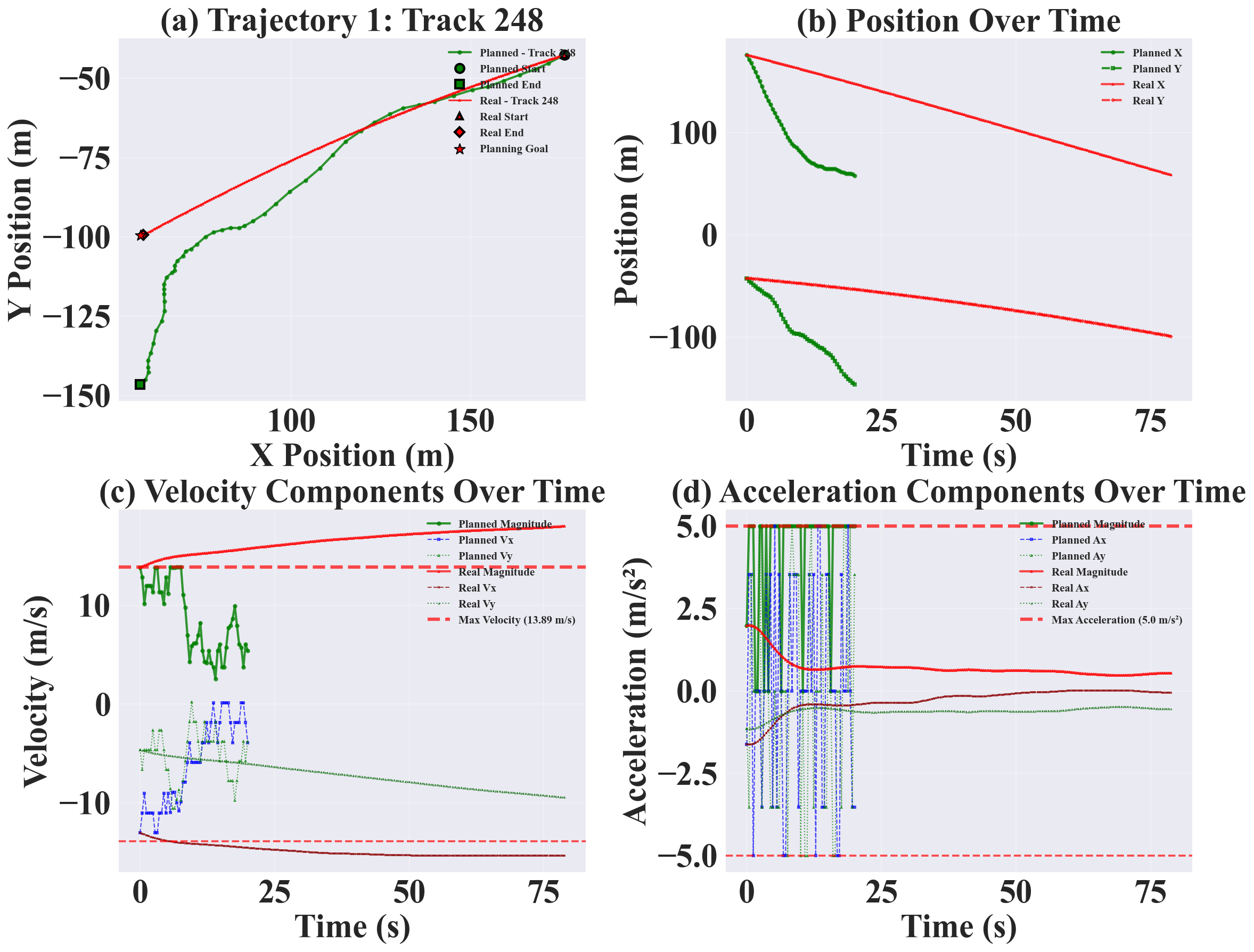}
    \caption{Illustration of constraint and trajectory (CPO).Reward Function: Progress to goal + Hard constraint penalties (velocity$\leq$13.89m/s, acceleration$\leq$5m/s², collision avoidance) + Soft constraint bonuses + Time penalty. Reward: -2394.62, Final Velocity: 2.27 m/s, Final Acceleration: 0.00 m/s², Distance Traveled: 83.28 m}
    \label{fig:constraint_diagram}
\end{figure}
DRIVE exhibits strong stability and robustness across five random seeds, with variance in success rate and violation rate below 0.1\%. The model converges stably within 30 episodes, demonstrating efficient learning dynamics.

\section*{Trajectory Generation Examples (MDP, BS)}

\begin{figure}[htbp]
    \centering
    \includegraphics[width=0.55\linewidth]{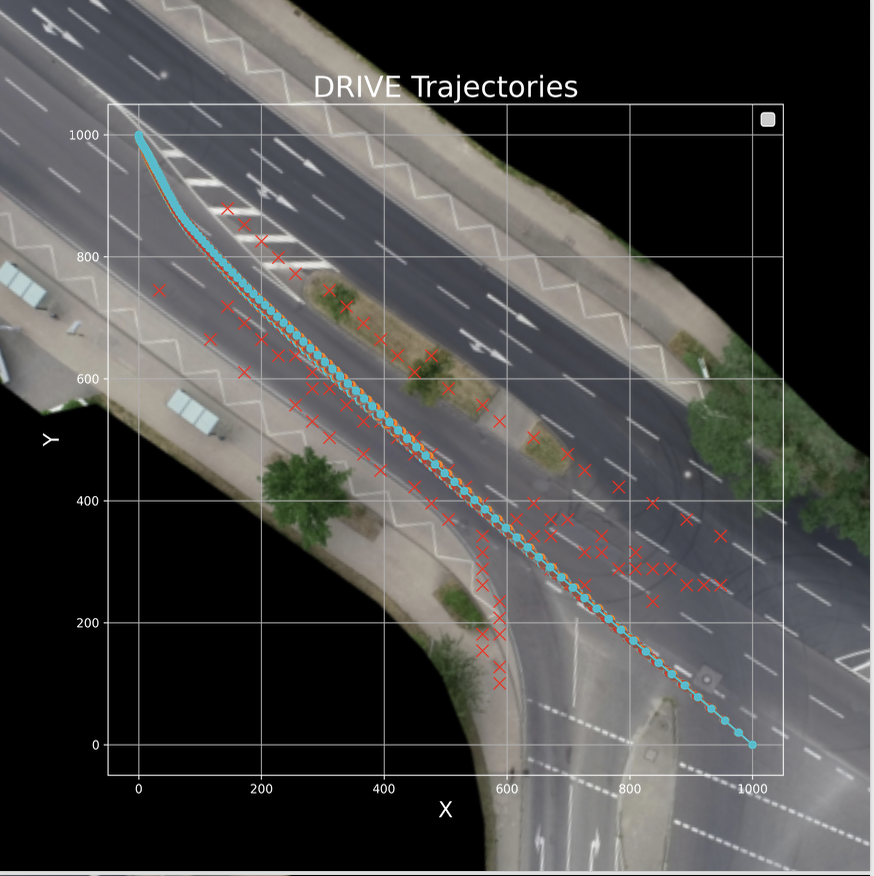}
    \caption{Beam Search at epoch 1. Trajectories are scattered and lack structure.}
    \label{beam_epoch1}

    \includegraphics[width=0.55\linewidth]{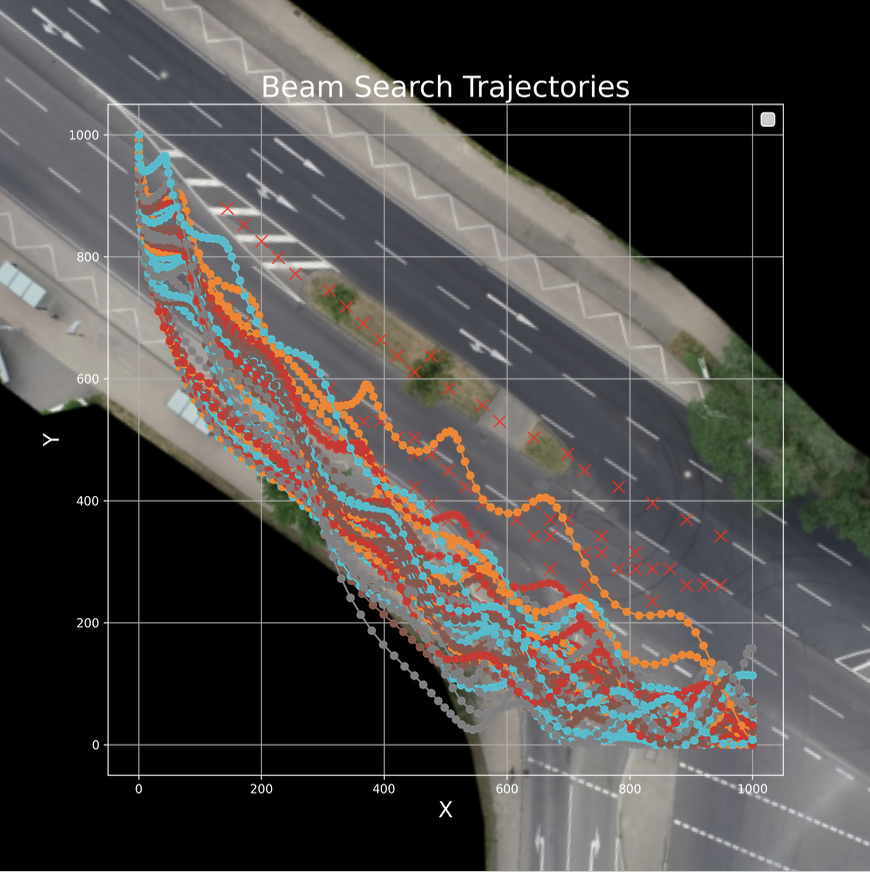}
    \caption{Beam Search at epoch 20. Trajectories begin to follow lane geometry.}
    \label{beam_epoch20}

    \includegraphics[width=0.55\linewidth]{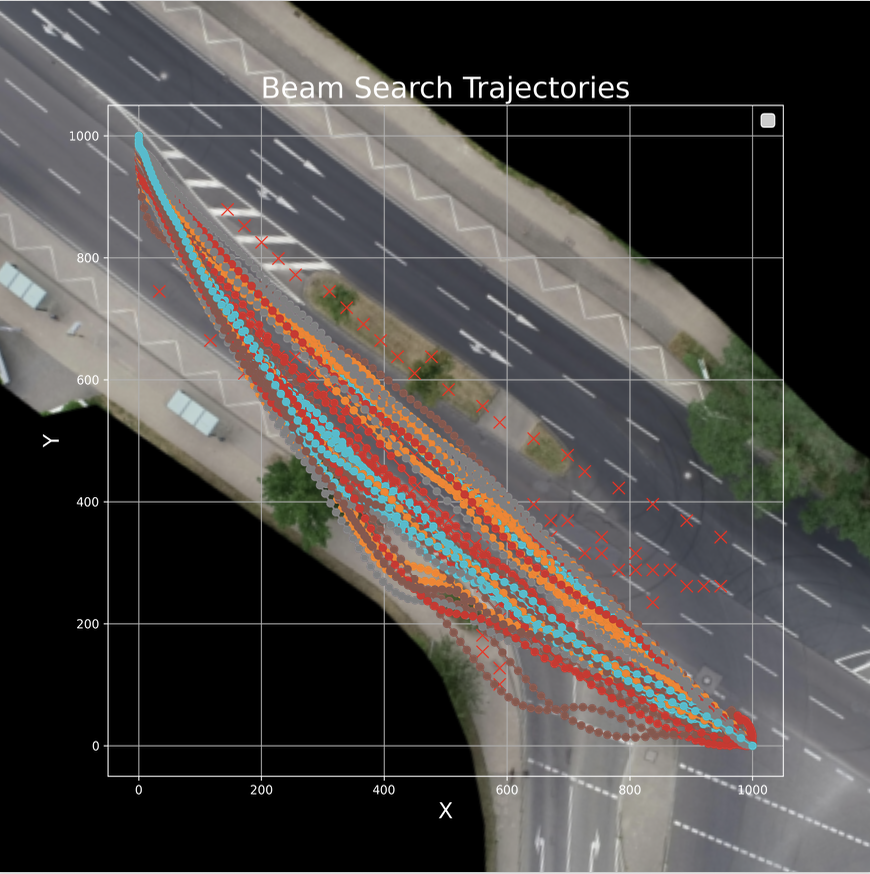}
    \caption{Beam Search at epoch 100. Final behavior becomes smooth and constraint-aware.}
    \label{beam_epoch100}
\end{figure}

Figures~\ref{beam_epoch1}--\ref{beam_epoch100} demonstrate the progression of Beam Search trajectory generation across training epochs. Initially, at epoch 1 (Figure~\ref{beam_epoch1}), the sampled trajectories are highly scattered and deviate significantly from road structure, indicating a lack of meaningful guidance. By epoch 20 (Figure~\ref{beam_epoch20}), the policy begins to align with the underlying lane geometry, although some variability remains. After 100 epochs (Figure~\ref{beam_epoch100}), the trajectories exhibit smooth, structured behavior that respects road boundaries and dynamic constraints, suggesting convergence to a constraint-aware and feasible motion policy.


\begin{figure*}[htbp]
    \centering
    \includegraphics[width=\textwidth]{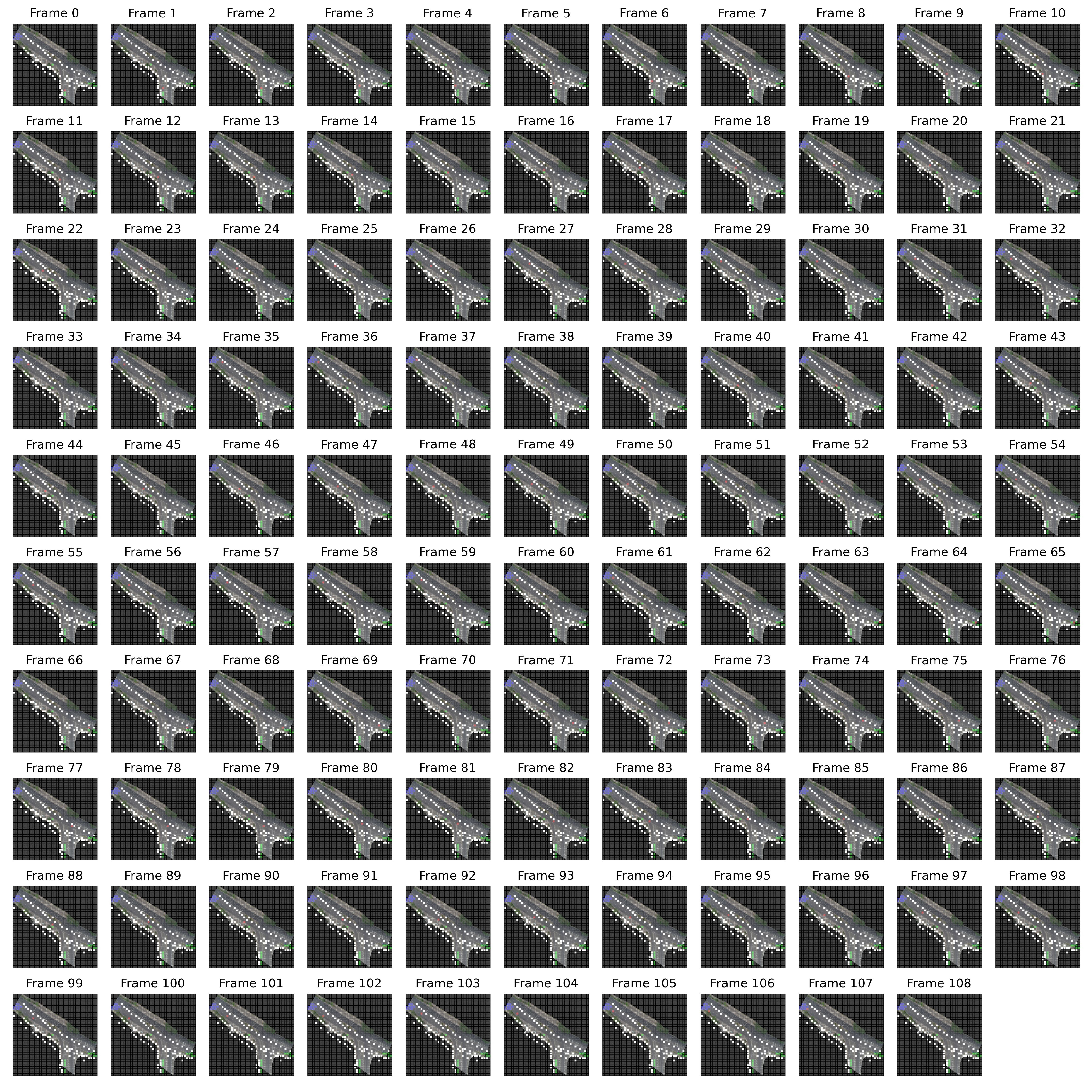}
    \caption{Trajectory generation via MDP policy (max frame = 108). Compared to dataset-based demonstration, the MDP policy introduces clear deviations from optimal constraint-following behavior, especially in tight curves or near lane changes.}
    \label{visualization_icl}
\end{figure*}

\end{document}